\documentclass[journal]{IEEEtran}
\usepackage[T1]{fontenc}% optional T1 font encoding
\usepackage{multirow,graphicx}
\usepackage{amsmath}
\usepackage{amssymb}
\usepackage{bbding}
\usepackage{threeparttable}
\usepackage{color}
\usepackage{url}
\usepackage{subfigure}
\usepackage{amssymb}
\usepackage{array}%
\usepackage{makecell}
\usepackage{algorithm}  
\usepackage{algorithmicx}  
\usepackage{algpseudocode}
\usepackage{bbding}
\usepackage{amssymb}
\usepackage{multirow}
\usepackage{amsmath,graphicx,multirow,bm}
\usepackage[switch,pagewise,columnwise]{lineno}
\usepackage[colorlinks,linkcolor=black]{hyperref}
\makeatletter
\let\NAT@parse\undefined
\makeatother
\usepackage{hyperref}
\ifCLASSINFOpdf
\else
\fi
\usepackage{amsmath}
\begin{document}
	%\linenumbers
	\title{
 Shuffle Mamba: State Space Models with Random Shuffle for Multi-Modal Image Fusion
 % Wavelet Empowered Pan-sharpening with Enhanced High-Frequency Details
 }
	
	\author{
		Ke Cao,
		Xuanhua He,\thanks{
    		 This work was supported by Anhui Provincial Natural Science Foundation under Grant 2408085MD090. Ke Cao and Xuanhua He contributed equally to this work; Corresponding authors: Man Zhou and Jie Zhang.

Ke Cao, Xuanhua He, and Tao Hu are with the University of Science and Technology of China, Hefei 230026, and the Institute of Intelligent Machines, and Hefei Institutes of Physical Science, Chinese Academy of Sciences, Hefei 230031, China(e-mail: caoke200820, hexuanhua, ht\_simon@mail.ustc.edu.cn);

% Ke Cao, Xuanhua He, and Tao Hu are with the Institute of Intelligent Machines, Hefei Institutes of Physical Science, Chinese Academy of Sciences, Hefei 230031, and the University of Science and Technology of China, Hefei 230026, China(e-mail: caoke200820, hexuanhua, ht\_simon@mail.ustc.edu.cn);

Chengjun Xie and Jie Zhang are with the Intelligent Agriculture Engineering Laboratory of Anhui Province, Institute of Intelligent Machines, and Hefei Institutes of Physical Science, Chinese Academy of Sciences, Hefei 230031, China (e-mail: cjxie, zhangjie@iim.ac.cn);

Man Zhou is with the University of Science and Technology of China, Hefei 230026, China (e-mail:manman@mail.ustc.edu.cn).
		}
            Tao Hu,
            Chengjun Xie,
            Man Zhou,
            Jie Zhang
		% <-this % stops a space% <-this %
	}

	\markboth{IEEE Transactions on Circuits and Systems for Video Technology}%
	{Shell \MakeLowercase{\textit\emph{et al.}}: Bare Demo of IEEEtran.cls for IEEE Communications Society Journals}

	\maketitle
 
	\begin{abstract}
Multi-modal image fusion integrates complementary information from different modalities to produce enhanced and informative images. Although State-Space Models, such as Mamba, are proficient in long-range modeling with linear complexity, most Mamba-based approaches use fixed scanning strategies, which can introduce biased prior information. To mitigate this issue, we propose a novel Bayesian-inspired scanning strategy called Random Shuffle, supplemented by a theoretically feasible inverse shuffle to maintain information coordination invariance, aiming to eliminate biases associated with fixed sequence scanning.  Based on this transformation pair, we customized the Shuffle Mamba Framework, penetrating modality-aware information representation and cross-modality information interaction across spatial and channel axes to ensure robust interaction and an unbiased global receptive field for multi-modal image fusion. Furthermore, we develop a testing methodology based on Monte-Carlo averaging to ensure the model's output aligns more closely with expected results. Extensive experiments across multiple multi-modal image fusion tasks demonstrate the effectiveness of our proposed method, yielding excellent fusion quality compared to state-of-the-art alternatives. The code is available at https://github.com/caoke-963/Shuffle-Mamba.
\end{abstract}
	
	
\IEEEpeerreviewmaketitle

\section{Introduction}
Multi-modal image fusion, a fundamental task in computer vision, involves extracting and integrating valuable information from images of the same scene captured by different imaging modalities~\cite{tcsvt_fusion1, tcsvt_fusion2}. This task aims to create a single composite image with a more comprehensive and informative representation, with typical applications including pan-sharpening and medical image fusion (MIF). In the context of pan-sharpening, satellites are limited by sensors, which can only capture low-resolution multi-spectral (LRMS) and panchromatic (PAN) images. Specifically, PAN images offer superior spatial details but limited spectral resolution, while MS images provide abundant spectral resolution but lack spatial clarity. By integrating the complementary information from both MS and PAN images into a composite representation, we can achieve an effective balance between spatial and spectral resolution. Analogously, in the realm of MIF, various imaging technologies capture distinct types of information. For instance, Computed Tomography (CT) images deliver detailed insights into bones and high-density tissues, while Magnetic Resonance Imaging (MRI) offers higher-resolution images with rich soft tissue details. 
In virtue of this complementary information from various modalities, MIF can overcome the limitations of single-modality images, resulting in a more comprehensive and detailed representation of modern medical diagnosis.
\begin{figure}[t]
% \captionsetup{type=figure}
\centering
\includegraphics[width=\linewidth]{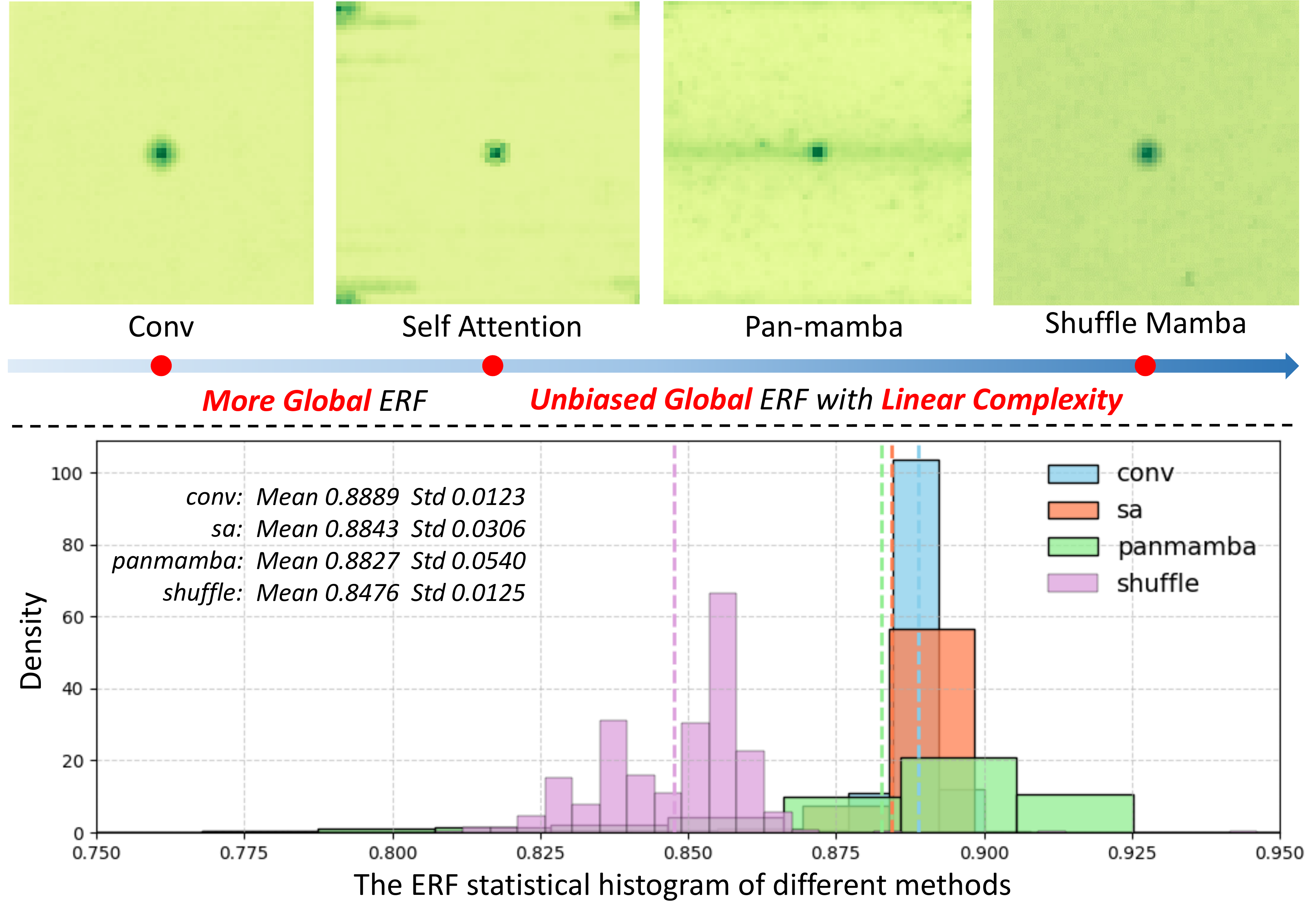}
\caption{\label{erf} Visualization of Effective Receptive Fields (ERFs) and ERF Intensity Distributions for Conv, Self-Attention, Pan-Mamba, and Our Method. The upper part displays ERF visualizations, where larger, darker regions indicate broader receptive fields. The lower part presents the histogram of ERF intensities, with the x-axis representing ERF strength (lower values correspond to darker colors and stronger ERF responses) and the y-axis denoting pixel probability density. Additionally, the mean and standard deviation of ERF intensities are reported to provide a comprehensive comparison.
}
% \vspace{-0.5em}
\end{figure}

In recent years, the prosperous advancement of deep neural networks (DNNs) has led to the development of numerous DNN-based multi-modal image fusion methods. In pan-sharpening, the pioneering work PNN~\cite{pnn} employed a simple three-layer neural network to achieve remarkable results that were previously deemed unattainable, highlighting the superior learning capabilities of deep learning. After that, increasingly complex and deeper architectures emerged~\cite{zhou2022sfiin, yang2023panflownet, zhang2024frequency}, delivering excellent visual performance. Despite these advancements, existing multi-modal fusion methods face common limitations.
Beyond improving fusion quality in standard scenarios, a growing trend in the field is to address image fusion in complex and adverse conditions. Recent studies have proposed unified frameworks to handle multiple complex scenes~\cite{li2025umcfuse} and developed large-scale benchmarks to facilitate all-weather multi-modality fusion~\cite{li2024allweather}. Furthermore, incorporating global and local text perception has proven effective for fusion in adverse weather~\cite{li2025awmweather}. These works highlight the increasing demand for robust perception capabilities in real-world applications.
Convolutional neural network (CNN)-based approaches often struggle to establish global receptive fields. While transformers address this issue through self-attention mechanisms, they introduce significant challenges related to quadratic computational complexity. Nowadays, structured state-space models have gained considerable attention for their computational efficiency and principled ability to model long-range dependencies. However, their selective scanning mechanism can introduce biased priors when processing 2D images. To overcome these challenges, it is reasonable for us to design a novel sequential scanning method with its application framework.

\textbf{Our Motivation.} Global modeling capabilities are essential in image restoration tasks because one of the core aspects of image restoration is to find useful information within the image to compensate for the missing data in the current patch. For a long time, CNN and ViT have been dominant architectures in computer vision, each with distinct advantages and limitations. CNNs are constrained by local receptive fields, which hinder their ability to model long-range dependencies. In contrast, ViT uses the self-attention mechanism to access a global receptive field but is burdened by quadratic computational complexity. Structured state-space models have recently demonstrated enhanced capabilities in capturing long-term dependencies in sequence data while maintaining linear-time complexity. In particular, Mamba has made significant strides in reducing inference latency and improving overall performance through selective state spaces and hardware-aware algorithms. With the introduction of VMamba~\cite {Vmamba} and Vision Mamba~\cite{VisionMamba}, there has been a growing interest in applying state-space models to visual tasks.

Most existing Mamba-based vision models adopt unidirectional or fixed scanning strategies to convert 2D spatial data into 1D sequences, which introduces fundamental limitations in receptive field modeling. Due to the sequential nature of state space models, earlier tokens typically benefit from broader receptive fields, while later tokens suffer from reduced context, leading to unbalanced global dependency modeling. Moreover, unlike the inherently causal structure of language, visual data is spatially non-causal and two-dimensional. Flattening image patches and scanning them sequentially disrupts spatial continuity, introduces orientation-specific biases, and hampers the modeling of local context. Prior studies, such as DgMamba~\cite{long2024dgmamba} and EAMamb~\cite{lin2025eamamba}, have shown that fixed scanning breaks adjacent pixel relations, while MambaIR~\cite{guo2024mambair} and MS-VMamba~\cite{shi2024msvmamba} demonstrate that such strategies lead to uneven receptive field distributions and long-range information decay. These issues persist even with multi-directional scanning, indicating that deterministic scanning orders inherently limit the spatial modeling capacity of Mamba-based frameworks.

To address these challenges, we propose a new sequence scanning method, termed Random Shuffle Scanning.
Figure~\ref{erf} visualizes the effective receptive fields (ERFs) of four representative approaches. Compared with convolutional models, self-attention achieves global perception but at the cost of quadratic computational complexity. Pan-Mamba achieves global modeling with linear complexity; however, its deterministic scanning order introduces structural bias, leading to an overemphasis on specific image patterns, such as horizontal stripes. In contrast, the proposed Shuffle Mamba establishes a balanced global receptive field with linear-time complexity, effectively removing fixed local priors and allowing the model to focus on task-relevant regions adaptively.
To further investigate the receptive field distribution, the lower part of Figure~\ref{erf} presents histograms of pixel-wise ERF intensities. The horizontal axis denotes gradient-based response intensity, while the vertical axis represents probability density. Compared with other methods, Shuffle Mamba exhibits a left-shifted distribution with a smaller mean value, corresponding to darker regions in the visualization and stronger activation responses. These results indicate that our randomized scanning promotes stronger global correlations and more uniform dependency modeling across spatial locations, effectively mitigating local bias and achieving theoretically unbiased global modeling.
For the sequentially input image patches, we first apply position encoding and then randomly shuffle the patches before processing them through the Mamba block for long-range dependency modeling. This random shuffle operation removes the deterministic correlation between local and global 2D dependencies in expectation, enabling the model to access an unbiased prior and achieve consistent global perception. Recognizing that spatial shuffling may disrupt semantic consistency, a corresponding inverse transformation is applied to restore the original order of input patches after Mamba processing. This information-preserving shuffle–inverse pair serves as the foundation for the three key modules of our framework: the Random Mamba Block, Random Channel Interactive Mamba Block, and Random Modal Interactive Mamba Block. Furthermore, inspired by Dropout, we employ Monte Carlo averaging during inference to approximate the expected output, ensuring that the final prediction aligns closely with the theoretical expectation.

Our contributions can be summarized as follows:
\begin{itemize}
    \item We design the Shuffle Mamba framework, where the random shuffle operation in the key component provides an expected unbiased global receptive field without increasing any parameters.
    \item We develop a specific strategy for training and testing this framework. During training, each input is independently scanned using a random shuffle operation. During testing, we use Monte Carlo averaging to estimate the output of each Mamba block.
    \item Extensive experiments on two prominent multi-modal image fusion tasks demonstrate that our method accomplishes excellent performance in both quantitative assessments and visual quality.
\end{itemize}
\section{Related Work}\label{s2}
\subsection{Pan-sharpening}
Traditional pan-sharpening methods can be categorized into three main groups: Component Substitution (CS)-based methods~\cite{GS,GFPCA,Brovey}, Multi-Resolution Analysis (MRA)-based methods~\cite{HPF,ATWT1999}, and Variational Optimization (VO)-based methods~\cite{fasbender2008bayesian,tv}. 
CS-based approaches extract the spatial details of the PAN image in the transform domain and inject them into the spatial components of the LRMS image. Conversely, MRA-based methods rely on a multi-resolution decomposition to derive the spatial information from the MS image and replace it with the content of the PAN image. VO-based models incorporate prior knowledge to formulate the fusion process and solve it iteratively. However, despite their diverse strategies, these traditional methods often experience significant spectral or spatial distortions due to inappropriate transformation schemes and ill-posed solution conditions.

Inspired by SRCNN~\cite{srcnn} in image super-resolution, PNN~\cite{pnn} adopted a simple model consisting of a three-layer convolutional neural network and achieved remarkable results, which has been seen as pioneering work for deep learning in pan-sharpening. This paved the way for many subsequent DL-based approaches with more complex structures. For instance, PanNet~\cite{pannet} conducts high-frequency filtering and residual connections to preserve detailed information and spectrum in the image. MSDCNN~\cite{msdcnn} incorporates a multi-scale and multi-depth network to extract various-level features. Meanwhile, SFINet~\cite{zhou2022sfiin} and MSDDN~\cite{he2023msddn} leverage the Fourier transform and specially designed dual-domain structures to simultaneously capture spatial and frequency domain features, enabling deeper learning of high-frequency information. Transformer architecture has also been introduced to pan-sharpening~\cite{zhou2022panformer}, which employs self-attention mechanisms to enhance fusion quality. Additionally, considering the ill-posed property of the pan-sharpening problem, PanFlowNet~\cite{yang2023panflownet} adopts a flow-based model to learn the conditional distribution of the fused image, which can generate multiple variations of HRMS images during inference. More recently, recognizing the limitations of previous methods in global information modeling and cross-modal information fusion, Pan-mamba~\cite{he2024panmamba} customizes the core components based on the state space model, yielding excellent results in feature extraction and fusion.
\subsection{Medical Image Fusion}
Medical image fusion (MIF) plays a crucial role in improving disease diagnosis by combining essential information from different image sources into a single fused image. Since single-modality images capture limited information, MIF has become a vital tool to integrate diverse image features, facilitating more accurate and efficient diagnoses. Early methods for MIF predominantly relied on techniques such as spatial domain processing~\cite{shahdoosti2012spatial}, transform domain approaches~\cite{li2020laplacian}, and sparse representations~\cite{veshki2020image}. However, these traditional methods often struggled to capture deep semantic information and required manually designed fusion rules tailored to specific scenarios, which limited their generalizability. In recent years, deep learning techniques have gained increasing attention in the field of MIF. Liu et al.~\cite{liu2017medical} were among the first to apply convolutional neural networks (CNNs) to MIF tasks, directly learning the mapping from source images to fused outputs. Building on this, Zero~\cite{lahoud2019Zero} utilized pre-trained CNNs to extract deep feature maps and adjust fusion weights by comparing these maps. To tackle the challenge of modeling global context in image fusion, MSRPAN~\cite{fu2021MSRPAN} introduced a multi-scale pyramid network, residual connections, and attention mechanisms. This approach achieved superior visual quality and quantitative results by controlling the energy ratio during feature fusion. Furthermore, CDDFuse~\cite{zhao2023cddfuse}  introduced a correlation-driven Transformer-CNN dual-branch feature decomposition network, effectively separating modality-specific and modality-shared features. Inspired by the mixture of experts (MoE) model, TC-MoA~\cite{zhu2024tcmoa}  utilized task-specific routing strategies and customized adapters, achieving excellent performance across various image fusion tasks within a unified framework.

\begin{figure*}[h]
    \centering
    \includegraphics[width=\textwidth]{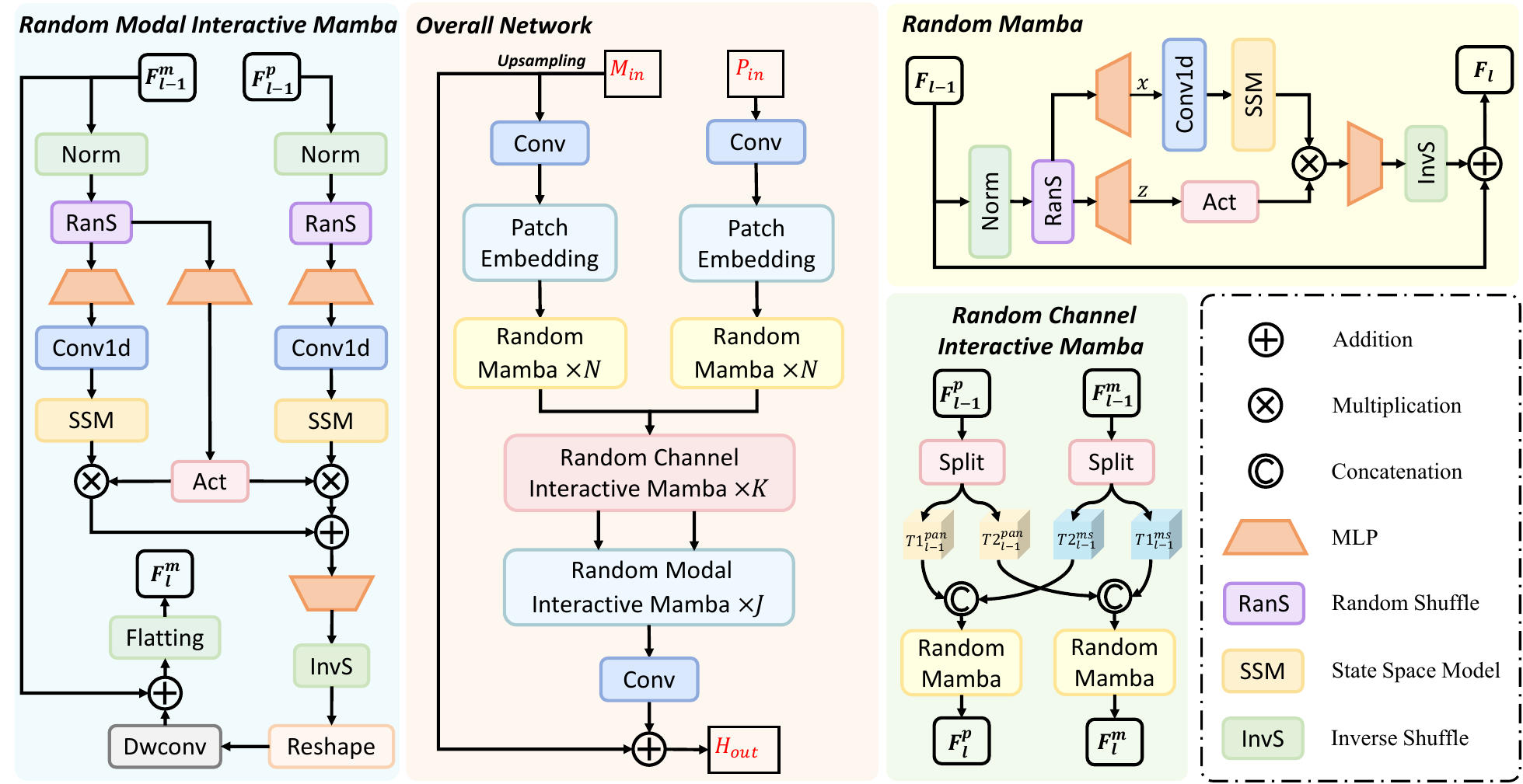}
    \caption{\label{method_mainfig}The architecture of the proposed Shuffle Mamba framework.}
    \label{method_mainfig}
\end{figure*}
\subsection{State Space Model}
The State Space Model (SSM), originating from control theory, has found extensive applications in deep learning due to its remarkable ability to model long-range dependencies. Initially, the S4~\cite{s4} model introduced the concept of the SSM, effectively reducing the computational and memory requirements associated with state representation while enabling global information modeling. Building upon S4, the S5~\cite{smith2022simplified} model features a MIMO structure and an efficient parallel scanning strategy to enhance performance without significantly increasing computational demands. Furthermore, the H3~\cite{h3} model further refines these methods, achieving competitive performance and efficiency comparable to the transformer in language modeling tasks.

Recently, Mamba~\cite{Mamba} has significantly improved inference speed and performance metrics through selective state spaces and hardware-aware algorithms. The introduction of Vmamba~\cite{Vmamba} and Vision Mamba~\cite{VisionMamba} has brought attention to the application of SSM in vision tasks. However, most existing SSM-based vision models~\cite{VisionMamba,ma2024umamba,zheng2024uvmamba} employ a fixed scanning strategy, which may introduce preconceptions, particularly in low-level vision tasks~\cite{xiao2023randomshuffle}. Specifically, this fixed order of selecting image patches can cause the model to gradually disregard previous input sequences while processing the current patch, thereby compromising its ability to model global information. To alleviate this challenge, Vmamba~\cite{Vmamba} introduced the CSM, which scans image pixels from various directions such as top-left, bottom-right, top-right, and bottom-left. Building on this idea, RSM~\cite{zhao2024rsmamba} introduced the OSSM to transform image patches into sequences in eight directions, thereby enhancing the network's ability to capture and model large-scale spatial features.
Additionally, LocalMamba~\cite{huang2024localmamba} applies the windowed selective scan to ensure a harmonious integration of global and local visual cues. RSMamba~\cite{chen2024rsmamba} incorporates dynamic multi-path activation mechanisms to model non-causal data. However, these methods do not simultaneously consider the integrity of the image's structure and the randomness of the pixels during scanning processing, and are essentially still special fixed strategies. 

To address this limitation, we introduced the Shuffle Mamba framework, which employs random shuffling and inverse operations to achieve a global receptive field without introducing expected biases. As a result, our approach not only advances pansharpening and medical image fusion but also demonstrates significant potential for a wider range of low-level vision tasks, including infrared and visible image fusion~\cite{bai2025refusion, bai2025task, jie2024tsjnet}, pan-denoising~\cite{xu2024pandenosing, xu2024hipandas}, and multi-modal UAV image fusion~\cite{jie2025fs}.

\section{Methods}\label{s3}
In this section, we introduce the foundational concepts of state-space models. Next, we describe the training strategy and module design underlying the proposed random shuffle framework. Finally, we detail the Monte Carlo averaging approach used during testing and discuss the design of the loss function tailored to different tasks.

\subsection{Preliminaries}
Inspired by continuous linear time-invariant (LTI) systems, SSMs exploit an implicit latent state $h(t)\in{\mathbf{R}^{N}}$to map a 1-D sequence $x(t)\in\mathbf{R}$ to $y(t)\in\mathbf{R}$. Specifically, SSMs can be mathematically expressed as an ordinary differential equation (ODE):
\begin{align}
    &{h}'(t)=\mathbf{A}h(t)+\mathbf{B}x(t),\\
    &y(t)=\mathbf{C}h(t).
\end{align}
Where $\mathbf{A}\in {\mathbf{R}^{N\times N}}$ is the evolution matrix, while $\mathbf{B}\in {\mathbf{R}^{N\times 1}}$ and $\mathbf{C}\in {\mathbf{R}^{1\times N}}$ serve as the projection parameters. However, solving these differential equations in a deep learning context can be challenging. The S4 and Mamba models propose introducing a timescale parameter $\mathbf{\Delta}$ to convert continuous parameters $\mathbf{A}$, $\mathbf{B}$ into their discrete counterparts $\bar{\mathbf{A}}$, $\bar{\mathbf{B}}$:
\begin{align}
  & {{h}_{\text{t}}}=\bar{\mathbf{A}}{{h}_{t-1}}+\bar{\mathbf{B}}{{x}_{t}}, \\ 
  & {{y}_{t}}=\mathbf{C}{{h}_{t}}, \\ 
  & \bar{\mathbf{A}}=\mathbf{\exp} \mathbf{(\Delta A)}, \\ 
  & \bar{\mathbf{B}}={{\mathbf{(\Delta A)}}^{-1}}(\exp \mathbf{(\Delta A)-I)}\cdot \mathbf{\Delta B}.
\end{align}
Finally, the output of the system can be attained through global convolution:
\begin{equation}\label{}
\bar{\mathbf{K}}=(\mathbf{C\bar{B}},\mathbf{C\bar{A}\bar{B}},...,\mathbf{C{{\bar{A}}^{L-1}}\bar{B}})
\end{equation}
where $\mathbf{L}$ represents the length of the sequence $x$, $\bar{\mathbf{K}}\in \mathbf{{R}^{L}}$ is a structured convolution kernel.

\begin{figure}[h]
% \captionsetup{type=figure}
\centering
\includegraphics[width=\linewidth]{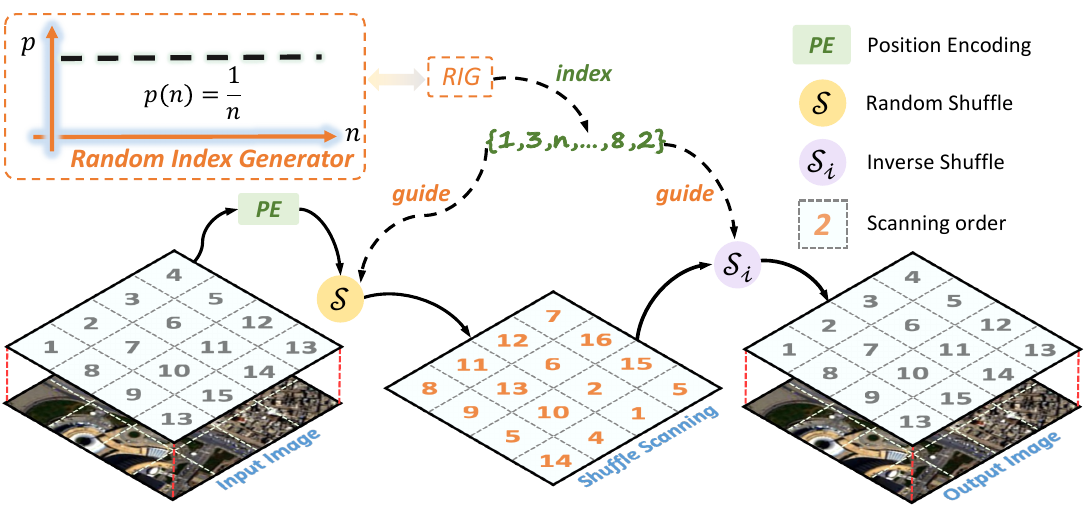}
\caption{\label{method_train}The Random Shuffle Scanning for training.
}
% \vspace{-0.5em}
\end{figure}
\begin{figure}[h]
% \captionsetup{type=figure}
\centering
\includegraphics[width=\linewidth]{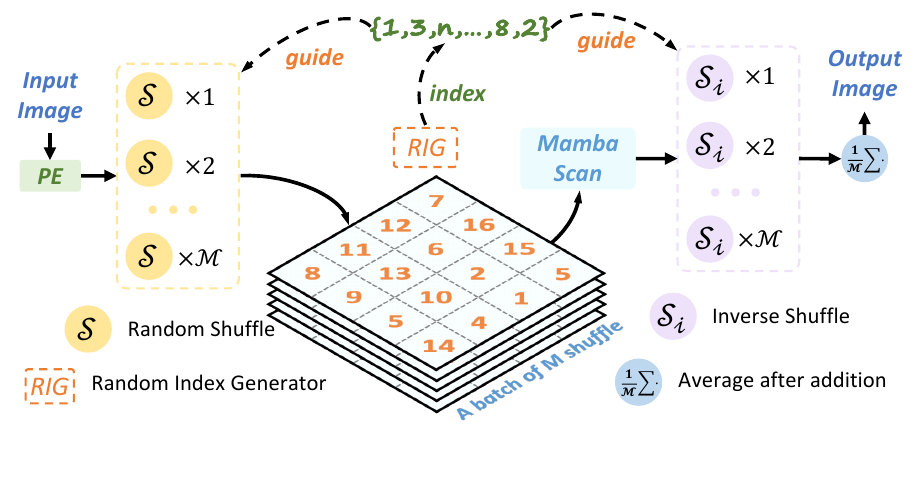}
\caption{\label{method_test}The Monte-Carlo averaging for testing.
}
% \vspace{-0.5em}
\end{figure}
\subsection{Network Architecture}
The proposed Shuffle Mamba framework consists of three functional components: the Random Mamba block (RM block), the Random Channel Interactive Mamba block (RCIM block), and the Random Modal Interactive Mamba block (RMIM block). The overall workflow is illustrated in Figure~\ref{method_mainfig}. Assuming that the input images with different modalities are ${\mathbf{M}_{up}}$ and ${\mathbf{P}_{in}}$, where ${\mathbf{M}_{up}} $ serve as the upsampled ${\mathbf{M}_{in}}$, we first use convolutional layers to project the images into the feature space. Given the limited receptive field of the convolutional layers, which makes capturing global features challenging, we perform patch embedding and send the resulting patches to the RM block for global feature extraction. This process yields the global modality-specific features $\mathbf{F}_{n}^{m}$ and $\mathbf{F}_{n}^{p}$, where $PE$ means the process of $PatchEmbed$,  $\phi$ denotes the initial convolution operation, $\psi$ corresponds to the Random Mamba module, and the subscripts indicate two branches and their sequential modules.
\begin{align}
    &\mathbf{F}_{0}^{m},\mathbf{F}_{0}^{p}=PE(\phi({\mathbf{M}_{up}})),PE(\phi({\mathbf{P}_{in}})),\\
    &\mathbf{F}_{n}^{m}={{\psi }_{1n}}\cdot \cdot \cdot ({{\psi }_{11}}({{\psi }_{10}}(\mathbf{F}_{0}^{m})),\\
    &\mathbf{F}_{n}^{p}={{\psi }_{2n}}\cdot \cdot \cdot ({{\psi }_{21}}({{\psi }_{20}}(\mathbf{F}_{0}^{p})).
\end{align}

The global modality-specific features are then sent to the RCIM block for simple channel information exchange, without introducing additional parameters. The exchanged features continue to be processed by their respective RM block to obtain $\mathbf{F}_{k}^{m}$ and $\mathbf{F}_{k}^{p}$. Next, we use the RMIM block to attain the $\mathbf{F}_{k+j}^{m}$ through the deep fusion of modality features $\mathbf{F}_{k}^{m}$ and $\mathbf{F}_{k}^{p}$. Thus, the final fused image ${\mathbf{H}_{out}}$ can be accessed by reshaping and residual connection:
\begin{align}
    &\mathbf{F}_{k+j}^{m}={{\theta }_{j}}\cdot \cdot \cdot ({{\theta }_{1}}({{\theta }_{0}}(\mathbf{F}_{k}^{m},\mathbf{F}_{k}^{p}),\mathbf{F}_{k}^{p}),\mathbf{F}_{k}^{p}),\\
    &{\mathbf{H}_{out}}=\phi (reshape(\mathbf{F}_{k+j}^{m}))+{\mathbf{M}_{up}}.
\end{align}

\subsection{Key Components}
\subsubsection{Random Shuffle Scanning}
Mamba was originally designed to adapt to the modeling of language sequences. We devised a Random Shuffle Scanning approach to access unbiased local and global dependencies in the 2D image while ensuring a global receptive field similar to self-attention. As shown in Figure~\ref{method_train}, for sequentially input 2D image patches, we first apply depth-wise convolution for position modeling. The image patches are then randomly shuffled and sent to the Mamba block for long-range dependency modeling. This strategy allows the Mamba block to simulate interactions between adjacent patches with equal probability, enabling the network to learn and model from an unbiased prior effectively. Additionally, since the relative positions of the patches are crucial for reconstructing semantic information, the output image must be accurately aligned with the input based on the inverse shuffle. For this reason, the random shuffle and its inverse operation constitute an information-lossless transformation pair.

\subsubsection{Random Mamba Block}
Based on this shuffle-inverse pair, we designed the Random Mamba Block. First, layer normalization is performed on the input feature ${\mathbf{F}_{l-1}}$  to obtain ${{\mathbf{F}'}_{l-1}}$ , which is then projected into $\mathbf{x}$ and $\mathbf{z}$ using random shuffle and multi-layer perceptrons (MLPs). In the first branch, $\mathbf{x}$ passes through 1-D convolution layers with SiLU activation to produce $\mathbf{x}'$. The SSM is then used to calculate the output $\mathbf{y}$. In the other branch, $\mathbf{z}$ is sent to the activation function to generate the gating for $\mathbf{y}$. Finally, we apply the inverse shuffle and residual connection to get the final output sequence ${\mathbf{F}_{l}}$.

\subsubsection{Random Channel Interactive Mamba Block}
In the RCIM block, we refer to the approach from~\cite{he2024panmamba}  to achieve lightweight feature interaction between different modalities. We use the split operation to divide modality features $\mathbf{F}_{k}^{m}$ and $\mathbf{F}_{k}^{p}$ into two halves based on the channel dimension, followed by complementary splicing. The exchanged features are then sent to their respective RM Blocks for processing. By repeating these steps, the global modality-specific features are initially fused.

\subsubsection{Random Modal Interactive Mamba Block}
Motivated by cross-attention, we designed the RMIM block for processing multi-modal image information. In this approach, we project the shuffled sequence features into a shared space and use a gating mechanism to learn complementary information under an unbiased prior, thereby reducing the interference of redundant features on the fusion results. We employ a process similar to RM Block to generate ${\mathbf{y}_{m}}$ and ${\mathbf{y}_{p}}$, and use the input $\mathbf{F}_{l-1}^{m}$ to generate the gating parameter $\mathbf{z}$ for dynamic adjustment of ${\mathbf{y}_{m}}$ and ${\mathbf{y}_{p}}$. The two outputs are then combined and projected, followed by inverse shuffle and reshape operations to align with the input sequence. Finally, the output $\mathbf{F}_{l}^{m}$ of the module can be obtained through depth-wise convolution and feature flattening.
\subsection{Testing with Monte Carlo averaging}
We incorporate stochastic factors into the random shuffle operation, necessitating the marginalization of these factors in the generation of the final fusion. However, the random shuffle method presents a theoretical challenge due to the exponentially large number of potential models, making precise averaging of their predictions infeasible. Drawing inspiration from dropouts~\cite{srivastava2014dropout}, we approximate the expected value of the entire model through layered expectations. Therefore, the computation of the random shuffle during testing can be expressed as follows, where $\mathbf{RM}=RandomMamba$:
\begin{equation}\label{}
\mathbf{R}{\mathbf{M}^{test}}(x)=\underset{\mathbf{\mathcal{S}}}{\mathop \mathbb{E}}\,[\mathbf{RM}(x,\mathbf{\mathcal{S}})]
\end{equation}
Estimating $\mathbf{R}{{\mathbf{M}}^{test}}$ based on the aforementioned equation requires enumerating all possible shuffle results, which imposes a significant computational burden. Therefore, we employ Monte Carlo averaging to estimate its expectation:
\begin{equation}\label{}
\mathbf{R}{\mathbf{M}^{test}}(x)\approx \frac{1}{M}\sum\limits_{i=1}^{M}{\mathbf{RM}(x,{{\mathbf{\mathcal{S}}}_{i}})}
\end{equation}
Specifically, the input image is independently shuffled ${M}$ times, and then the ${M}$ outputs of $\mathbf{RM}$ are calculated. The mean of these outputs is computed to obtain the final estimate. When ${M}\to \infty $ is present, the Monte Carlo estimator closely approximates the true mean. Figure~\ref{method_test} illustrates the test process we designed, which significantly reduces the actual testing time by incorporating multiple identical inputs in a mini-batch and utilizing GPUs for parallel computation.
\begin{table*}[htpb]
\centering
\normalsize
\renewcommand{\arraystretch}{1.1}
\renewcommand{\tabcolsep}{3pt}
\caption{Quantitative comparison of pan-sharpening task on three datasets. The best results are highlighted in \textbf{bold}. The second-best results are highlighted in \underline{underline}. $\uparrow$ indicates that the larger the value, the better the performance, and $\downarrow$ indicates that the smaller the value, the better the performance.}\label{table_pansharepening}
\resizebox{\linewidth}{!}
{
\begin{tabular}{c|cccc|cccc|cccc}
\hline
& \multicolumn{4}{c|}{WorldView-II}                                                      & \multicolumn{4}{c|}{Gaofen-2}                                                          & \multicolumn{4}{c}{WorldView-III}    \\ \cline{2-13} 
\multirow{-2}{*}{Method} 
& PSNR$\uparrow$      & SSIM$\uparrow$     & SAM$\downarrow$        & ERGAS$\downarrow$ & PSNR$\uparrow$      & SSIM$\uparrow$     & SAM$\downarrow$        & ERGAS$\downarrow$  & PSNR$\uparrow$      & SSIM$\uparrow$    & SAM$\downarrow$         & ERGAS$\downarrow$  \\ \hline
SFIM                     
& 34.1297           & 0.8975          & 0.0439           & 2.3449  
& 36.9060           & 0.8882          & 0.0318           & 1.7398 
& 21.8212           & 0.5457          & 0.1208           & 8.9730                        \\
Brovey                   
& 35.8646           & 0.9216          & 0.0403           & 1.8238
& 37.2260           & 0.9034          & 0.0309           & 1.6736    
& 22.5060           & 0.5466          & 0.1159           & 8.2331                        \\
IHS                      
& 35.6376           & 0.9176          & 0.0423           & 1.8774
& 37.7974           & 0.9026          & 0.0218           & 1.3720                        & 22.5608           & 0.5470          & 0.1217           & 8.2433                        \\
GS                       
& 35.2962           & 0.9027          & 0.0461           & 2.0278                        
& 38.1754           & 0.9100          & 0.0243           & 1.5336                        
& 22.5579           & 0.5354          & 0.1266           & 8.3616                        \\
GFPCA                    
& 34.5581           & 0.9038          & 0.0488           & 2.1411                        
& 37.9443           & 0.9204          & 0.0314           & 1.5604                        
& 22.3344           & 0.4826          & 0.1294           & 8.3964       \\ 
SFNLR                    
& 33.8636           & 0.8957          & 0.0481           & 2.2415              
& 36.2367           & 0.8850          & 0.0363           & 1.8611                      
& 21.7647           & 0.5636          & 0.1494           & 9.0962                     \\ 
LRTCFPan                    
& 34.7756                        & 0.9112                        & 0.0426                        & 2.0075                      
& 36.9253                        & 0.8946                        & 0.0332                        & 1.7060                      
& 22.1574                        & 0.5735                        & 0.1380                        & 8.6796          
\\ \hline
MSDCNN                   
& 41.3355                        & 0.9664                        & 0.0242                        & 0.9940                        
& 45.6847                        & 0.9827                        & 0.0135                        & 0.6389                        
& 30.3038                        & 0.9184                        & 0.0782                        & 3.1884                        
\\
% SRPPNN                   
% & 41.4538                        & 0.9679                        & 0.0233                        & 0.9899                        
% & 47.1998                        & 0.9877                        & 0.0106                        & 0.5586                        
% & 30.4346                        & 0.9202                        & 0.0770                        & 3.1553                       
% \\
INNformer            
& 41.6903                        & 0.9704                        & 0.0227                        & 0.9514                        
& 47.3528                        & 0.9893                        & 0.0102                        & 0.5479                        
& 30.5365                        & 0.9225                        & 0.0747                        & 3.1142                        
\\
SFINet            
& 41.7244                        & 0.9725                       & 0.0220                      & 0.9506                       
& 47.4712                        & \underline{0.9901}                        & 0.0102                      & 0.5479                       
& 30.5901                        & 0.9236                        & 0.0741                       & 3.0798                        
\\
MSDDN          
& 41.8435                        & 0.9711                        & 0.0222                        & 0.9478                       
& 47.4101                        & 0.9895 & 0.0101 &0.5414                      
& 30.8645                       & 0.9258                       & 0.0757                        & 2.9581                        
\\
PanFlowNet       
& 41.8548                       & 
0.9712                        & 0.0224                        & 0.9335                        
& 47.2533                        & 0.9884                        & 0.0103                        & 0.5512                        
& 30.4873                        & 0.9221                        & 0.0751                        & 2.9531                       
\\ 

FAME   
& 42.0262                       & 
0.9723                        & 0.0215                        & 0.9172                        
& \underline {47.6721}                        & 0.9898                        & \underline{0.0098}                        & 0.5542                        
& 30.9903                        & 0.9287                        & \underline{0.0697}                        & 2.9531                       
\\ 
DISPNet     
& 41.8768                       & 
0.9702                        & 0.0221                        & 0.9157                        
& 47.4529                        & 0.9898                        & 0.0111                        & 0.5532                        
& 30.0426                        & 0.9153                        & 0.0776                        & 3.2620                       
\\ 
Pan-mamba                 
& \underline {42.2354} & \underline {0.9729} & \underline {0.0212} & \underline {0.8975} 
& 47.6453 & 0.9894              & 0.0103              & \underline {0.5286} 
& \underline {31.1740} & \underline {0.9302} & 0.0698 & \underline {2.8910} 
\\ 
ARConv             
& 41.8281 & 0.9704 & 0.0222 & 0.9395 
& 47.3699 & 0.9887              & 0.0102              & 0.5492 
& 30.8156 & 0.9251 & 0.0717 & 3.0114

\\ \hline
Ours   
& \textbf {42.3428}   & \textbf {0.9734}  & \textbf {0.0208}  & \textbf {0.8840}
& \textbf {47.9180}   & \textbf {0.9903}  & \textbf {0.0097}  & \textbf {0.5109} 
& \textbf {31.4005}   & \textbf {0.9327}  & \textbf {0.0676}  & \textbf {2.8098} 
\\ \hline
\end{tabular}
}
\end{table*}
\subsection{Loss Function}
In accordance with established norms within this area, the loss function used in our model for pan-sharpening is the L1 loss. Assume the two input images of the model be represented as ${\mathbf{M}_{in}}$ and ${\mathbf{P}_{in}}$, with ${\mathbf{M}_{up}}$ obtained by upsampling ${\mathbf{M}_{in}}$.
The model output is denoted as ${\mathbf{H}_{out}}$, and the corresponding ground truth or basic reference features (if available) are represented as ${\mathbf{H}_{gt}}$, the loss of pan-sharpening ${\mathbf{L}_{pan}}$ can be articulated as follows:

\begin{equation}\label{}
\mathbf{\mathcal{L}_{pan}}={{\left\| {\mathbf{H}_{gt}}-{\mathbf{H}_{out}} \right\|}_{1}}
\end{equation}

In the MIF task, we utilize two input images and the fused image to compute a composite loss function ${\mathbf{L}_{fus}}$ comprising intensity loss ${\mathbf{L}_{1}}$, structural similarity index (SSIM) loss ${\mathbf{L}_{ssim}}$, and gradient loss ${\mathbf{L}_{grad}}$. 
\begin{align}
  & {\mathbf{L}_{1}}={{\left\| {\mathbf{H}_{out}}-\max ({\mathbf{M}_{up}},{\mathbf{P}_{in}}) \right\|}_{1}} \\ 
  & {\mathbf{L}_{ssim}}=0.5\times \mathbf{s}({\mathbf{H}_{out}},{{M}_{up}})+0.5\times \mathbf{s}({\mathbf{H}_{out}},{{P}_{in}}) \\ 
  & gra{{d}_{join}}=\max (\mathbf{sobel}({\mathbf{M}_{up}}),\mathbf{sobel}({\mathbf{P}_{in}}))\\ 
  & {\mathbf{L}_{grad}}={{\left\| \mathbf{sobel}({\mathbf{H}_{out}})-gra{{d}_{join}} \right\|}_{1}} \\ 
  & {\mathbf{L}_{fus}}=20{\mathbf{L}_{1}}+50{\mathbf{L}_{ssim}}+100{\mathbf{L}_{grad}}
\end{align}

Here, $\mathbf{s}$ denotes the structural similarity index (SSIM) calculation operation on the image, while $\mathbf{sobel}$ represents the Sobel convolution operation.

\begin{table*}[!htb]
\normalsize
\centering
\caption{\label{table_fwv2}
Evaluation of our method on real-world full-resolution scenes from the WV2 dataset. \textbf{Bold} and \underline{underline} show the best and second-best values, respectively.}
\resizebox{\linewidth}{!}{
\begin{tabular}{c|cccccccccccc}
\hline
Metric &SFIM & Brovey & GFPCA & INNformer &SFINet &PanFlowNet &FAME &DISPNet &Pan-mamba &ARConv &Ours\\
\hline
$D_{\lambda}$ $\downarrow$ & 0.1403 & 0.1026 & 0.1139  &0.0995 &0.1034&0.0966 &0.0951 & \underline{0.0944} &0.0966 &0.0963 &\textbf{{0.0941}} \\
$D_{S}$ $\downarrow$  & 0.1320 & 0.1409 & 0.1535 &0.1305&0.1305&0.1274 &\textbf{0.1263} & \underline{0.1264} &0.1272 &0.1276 &0.1266 \\
QNR $\uparrow$ & 0.7826 & 0.7728 & 0.7532 &0.7858 &0.7827&0.7910 & 0.7933 &\underline{0.7938} &0.7911 &0.7910 &\textbf{0.7939}\\ \hline
\end{tabular}
}
\end{table*}

\begin{figure*}[!htb]
    \centering
    \includegraphics[width=\textwidth]{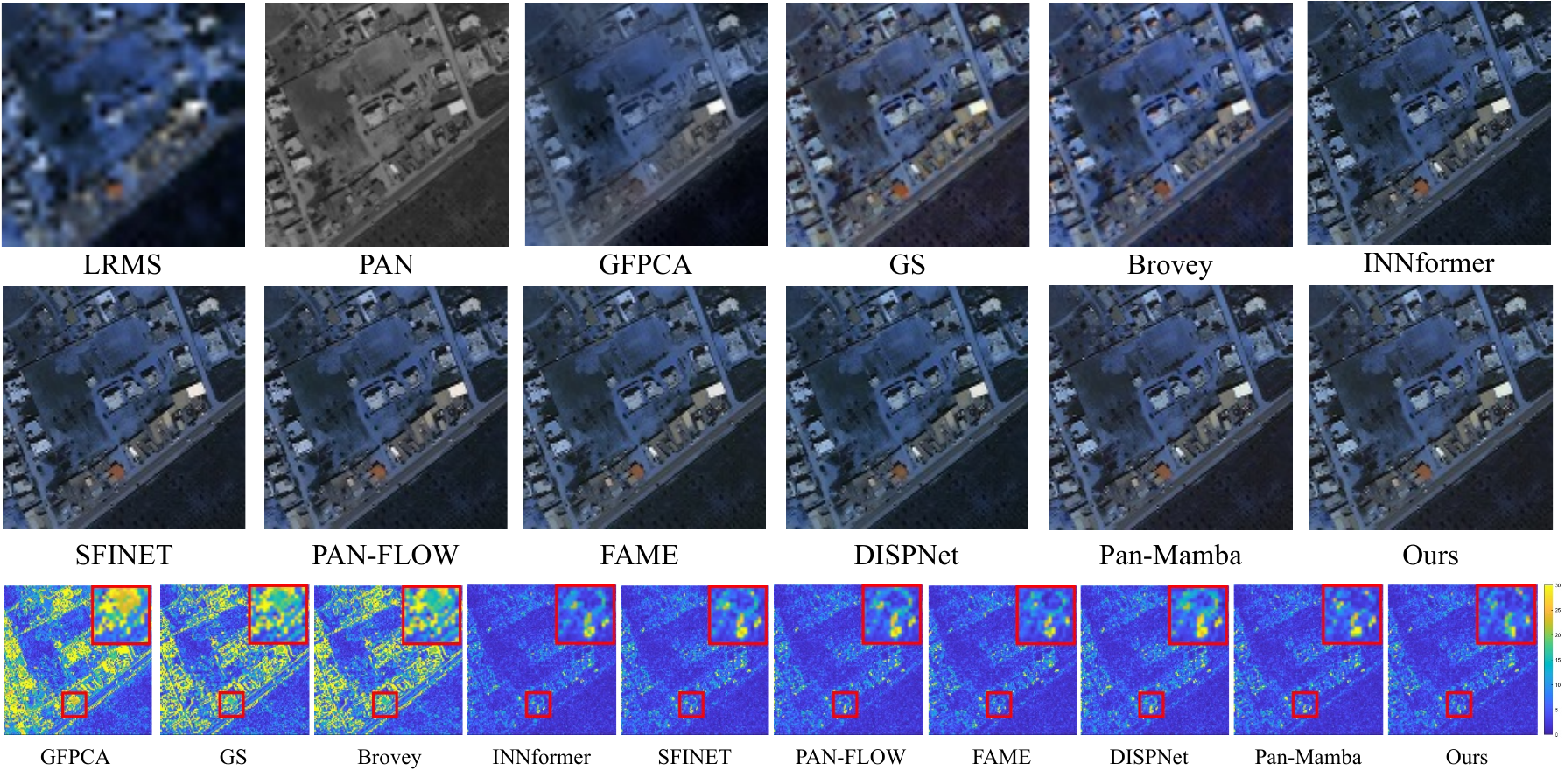}
    \caption{\label{result_WV3}Comparative visual experiments of several methods on WV3 datasets}
    \label{result_WV3}
\end{figure*}

\begin{figure*}[!htb]
    \centering
    \includegraphics[width=\textwidth]{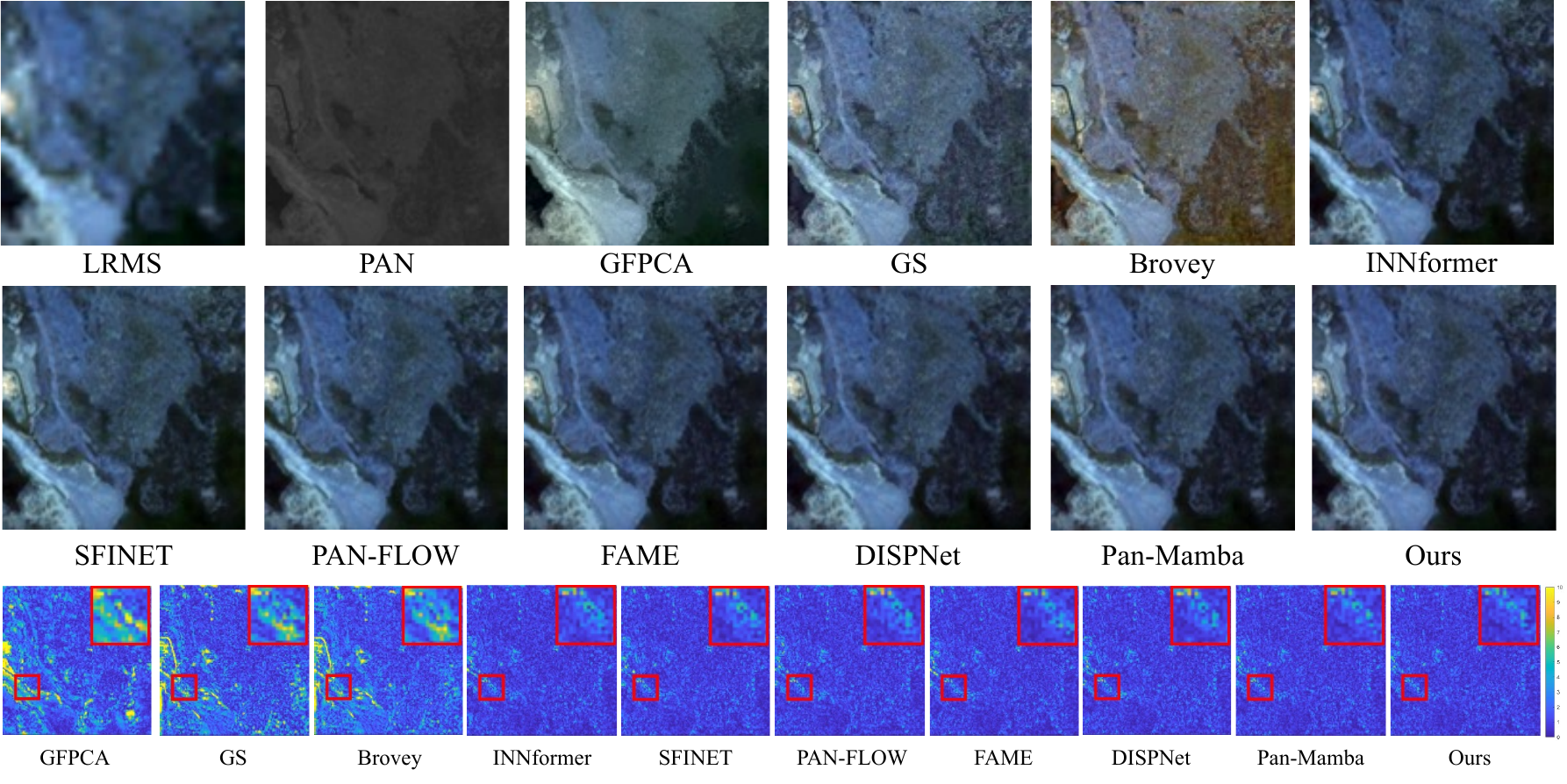}
    \caption{\label{result_WV2}Comparative visual experiments of several methods on WV2 datasets}
    \label{result_WV2}
\end{figure*}

\begin{figure*}[!htb]
    \centering
    \includegraphics[width=\textwidth]{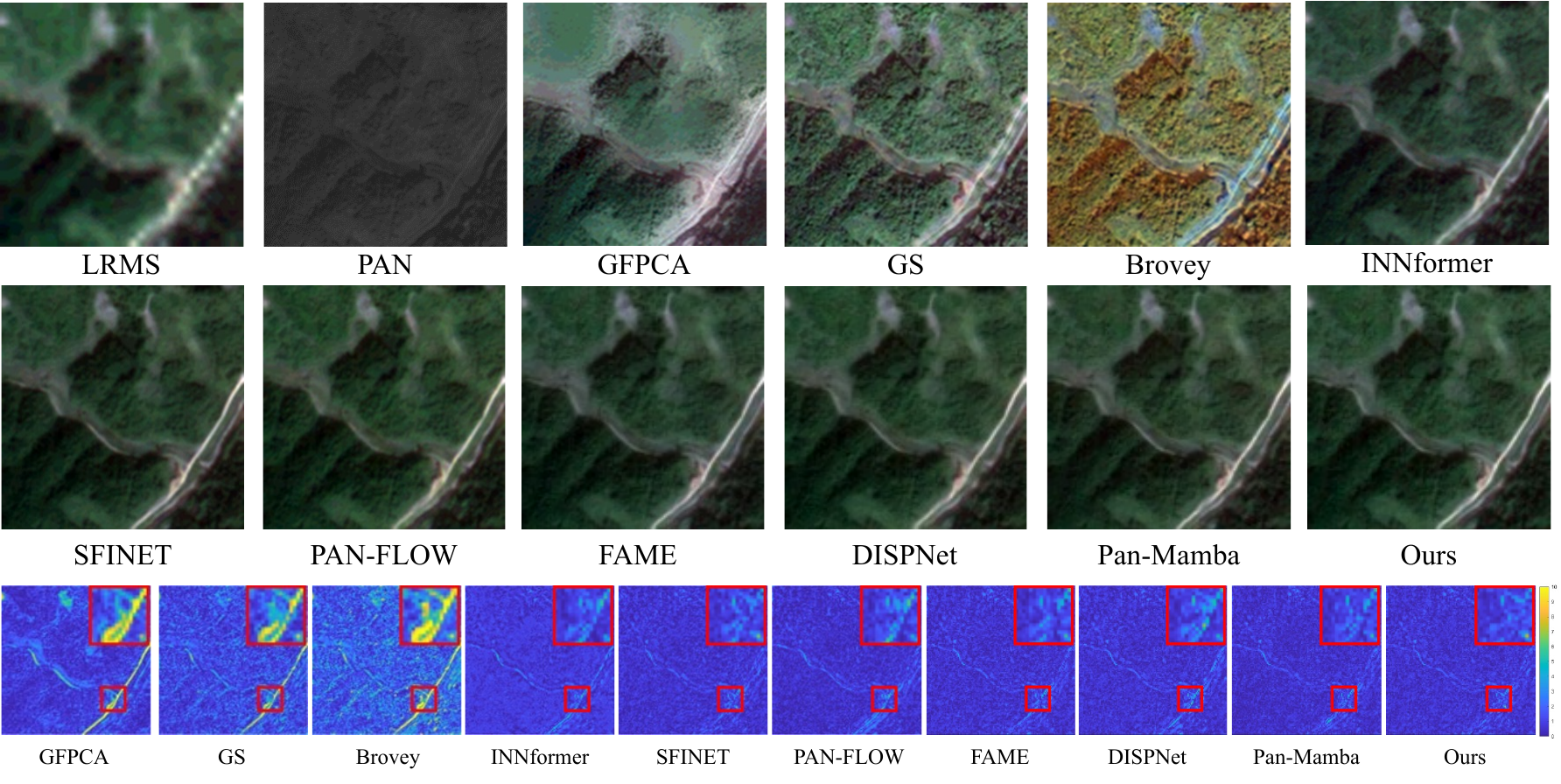}
    \caption{\label{result_GF2}Comparative visual experiments of several methods on GF2 datasets}
    \label{result_GF2}
\end{figure*}

\begin{figure*}[htpb]
    \centering
    \includegraphics[width=\textwidth]{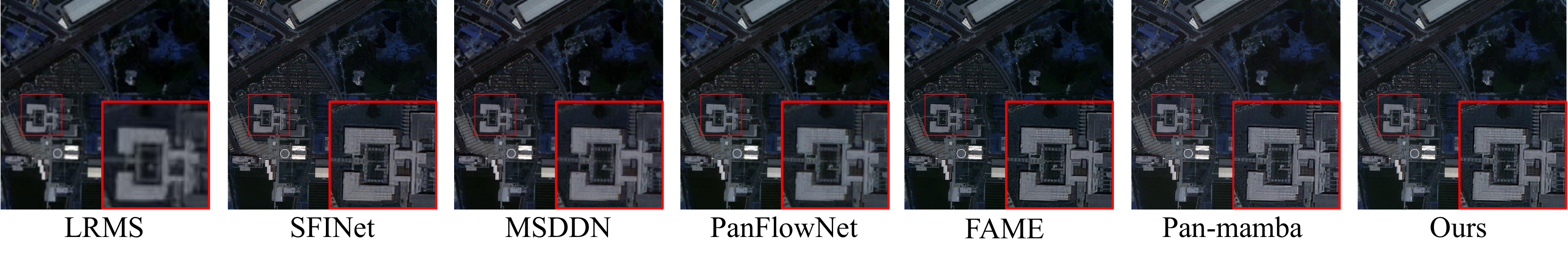}
    \caption{\label{result_full}The result of our approach on full-resolution WV2 dataset.}
    \label{result_full}
\end{figure*}

\section{Experiments}\label{s4}
\subsection{Datasets and Benchmark}
For the pan-sharpening task, we utilized datasets from WorldView-II (WV2), GaoFen2 (GF2), and WorldView-III (WV3), encompassing a diverse range of urban and natural scenes. Training samples were prepared using the Wald protocol~\cite{wald_gt}, which is widely adopted when the ground truth is inaccessible. To evaluate the effectiveness of our proposed method, we conducted a comprehensive comparison against both traditional and deep learning-based approaches. The baseline methods included classical algorithms such as SFIM~\cite{SFIM}, Brovey~\cite{Brovey}, GS~\cite{GS}, IHS~\cite{IHS}, GFPCA~\cite{GFPCA},  SFNLR~\cite{xiao2023variational} and LRTCFPan~\cite{wu2023lrtcfpan},as well as state-of-the-art deep learning techniques, including MSDCNN~\cite{msdcnn}, 
% SRPPNN~\cite{srppnn}, 
INNformer~\cite{zhou2022innformer}, SFINet~\cite{zhou2022sfiin}, MSDDN~\cite{he2023msddn}, PanFlowNet~\cite{yang2023panflownet}, FAME~\cite{he2024frequencymoe}, DISPNet~\cite{DISPNet2024}, Pan-Mamba~\cite{he2024panmamba}, and ARConv~\cite{wang2025arconv}.
\begin{table*}[!htb]
    \centering
    \small
    \renewcommand{\arraystretch}{1.1}
\renewcommand{\tabcolsep}{3pt}
\caption{Evaluation of parameter count and GFLOPs.}
    \label{table_flops}
\begin{tabular}{c|ccccccccc}
\hline
Methods   & SRPPNN & INNformer & MSDDN  & PanFlowNet & FAME & DISPNet  & Panmamba & ARConv & Ours   \\ \hline
Params(M) & 1.7114 & 0.0706 & 0.3185  & 0.0873  & 0.6286 & 0.4433 & 0.1827   & 4.4147 & 0.2077 \\
GFLOPs(G)  & 21.1059 & 1.3079 & 2.5085 & 5.7126  & 10.1943 & 7.7659 & 3.0088  & 14.0833  & 3.4021 \\
Infer Time(ms)  & 6.7071 & 11.9172 & 7.4857 & 5.3810  & 12.7889 & 11.5221 & 19.0202  & 63.3764  & 24.4713 

\\ \hline
\end{tabular}
\end{table*}

\begin{table*}[!h]
\centering
\normalsize
\renewcommand{\arraystretch}{1.0}
\renewcommand{\tabcolsep}{3pt}
\caption{ Quantitative comparison of MIF task on three datasets. \textbf{bold} and \underline{underline} show the best and second-best values. \label{table_MIF}}
\resizebox{\linewidth}{!}
{
\begin{tabular}{c|cccc|cccc|cccc}
\hline
& \multicolumn{4}{c|}{PET}                                                      
& \multicolumn{4}{c|}{CT}                                                          
& \multicolumn{4}{c}{SPECT}    \\ \cline{2-13} 
\multirow{-2}{*}{Method} 
& SCD$\uparrow$      & VIF$\uparrow$     & Qabf$\uparrow$        & SSIM$\uparrow$ 
& SCD$\uparrow$      & VIF$\uparrow$     & Qabf$\uparrow$        & SSIM$\uparrow$  
& SCD$\uparrow$      & VIF$\uparrow$     & Qabf$\uparrow$        & SSIM$\uparrow$  
\\ \hline
EMFusion                     
& 0.943           & 0.685          & \underline{0.783}           & 0.611  
& 1.190           & \textbf{0.552}          & 0.475           & 0.633 
& 0.885           & 0.665          & 0.692           & 0.606                        \\
MSRPAN                   
& 1.017           & 0.581          & \textbf{0.799}           & 0.591
& 1.319           & 0.436          & 0.455           & 0.631    
& 0.960           & 0.525          & 0.560           & 0.577                        \\
SwinFusion                      
& \textbf{1.642}           & 0.703          & 0.683           & 0.363
& 1.537           & 0.522          & 0.545           & 0.290       
& \textbf{1.678}           & 0.744          & 0.720           & 0.342                        \\
Zero                    
& 0.950           & 0.635          & 0.774           & 0.581
& 0.768           & 0.320          & 0.582           & 0.600
& 1.046           & 0.582          & 0.681           & 0.590            
\\
U2Fusion                   
& 0.947           & 0.460          & 0.292           & 0.247            
& 0.309           & 0.074          & 0.489           & 0.021            
& 0.865           & 0.419          & 0.696           & 0.240                 
\\
CDDFuse                   
& 1.481           & 0.650          & 0.765           & 0.614                        
& \underline{1.589}           & 0.526          & 0.530           & 0.612                        
& 0.995           & \underline{0.786}          & 0.719           & 0.585   
\\
PSLPT                     
& 0.888           & 0.548          & 0.373           & 0.408  
& 0.675           & 0.502          & 0.432           & 0.405 
& 0.850           & 0.359          & 0.325           & 0.467   
\\
Fusionmamba                     
& 1.399           & \underline{0.791}          & 0.736           & 0.618  
& 1.532           & 0.538          & \underline{0.587}           & \underline{0.664} 
& 1.368           & 0.761          & \underline{0.732}           & \underline{0.619} 
\\   
TC-MoA                   
& 1.387           & 0.540          & 0.615           & 0.563  
& 1.584           & 0.451          & 0.503           & 0.414 
& 1.487           & 0.554          & 0.594           & 0.557
\\ 
PRRGAN               
& 1.405           & 0.641          & 0.696           & 0.607  
& 1.477           & 0.502          & 0.497           & 0.563 
& 1.109           & 0.682          & 0.701           & 0.589
\\ 
TFS-Diff                   
& 1.489           & 0.665          & 0.758           & \underline{0.621}  
& \textbf{1.591}           & 0.530          & 0.547           & 0.658 
& \underline{1.549}           & 0.743          & 0.684           & 0.611
\\ 
\hline
Ours                     
& \underline {1.491}           & \textbf{0.797}          & 0.741           & \textbf{0.628}
& 1.580 &  \underline{0.546}       &  \textbf{0.592}        &  \textbf{0.665} 
&  1.470 &  \textbf{0.820} &  \textbf{0.747} &  \textbf {0.620} \\ \hline
\end{tabular}
}
\end{table*}

\begin{table*}[!h]
\centering
\normalsize
\renewcommand{\arraystretch}{1.0}
\renewcommand{\tabcolsep}{3pt}
\caption{Quantitative comparison of IVIF task on three datasets. \textbf{bold} and \underline{underline} show the best and second-best values. \label{table_IVIF}}
\resizebox{\linewidth}{!}{
\begin{tabular}{c|cccc|cccc|cccc}
\hline
& \multicolumn{4}{c|}{MSRS}                                                 
& \multicolumn{4}{c|}{RoadScene}                                            
& \multicolumn{4}{c}{M$^{3}$FD}    \\ \cline{2-13} 
\multirow{-2}{*}{Method} 
& SCD$\uparrow$      & VIF$\uparrow$     & Qabf$\uparrow$        & SSIM$\uparrow$ 
& SCD$\uparrow$      & VIF$\uparrow$     & Qabf$\uparrow$        & SSIM$\uparrow$  
& SCD$\uparrow$      & VIF$\uparrow$     & Qabf$\uparrow$        & SSIM$\uparrow$  
\\ \hline

SemLA                      
& 1.392 & 0.608 & 0.290 & 0.498
& 1.269 & 0.564 & 0.415 & 0.518   
& 1.495 & 0.542 & 0.363 & 0.473  
\\
U2Fusion                   
& 1.258 & 0.512 & 0.391 & 0.440     
& 1.605 & 0.564 & 0.506 & \underline{0.546}        
& \underline{1.753} & 0.673 & 0.578 & 0.463          
\\
SwinFusion                   
& 1.647 & 0.825 & 0.558 & 0.504                  
& 1.576 & 0.614 & 0.450 & 0.534                   
& 1.588 & 0.746 & 0.616 & 0.492
\\
CDDFuse                    
& 1.549 & 0.819 & 0.548 & 0.459
& 1.707 & 0.610 & 0.450 & 0.515
& 1.673 & 0.802 & 0.613 & 0.460
\\
PSLPT                     
& 1.374 & 0.753 & 0.553 & 0.501
& 1.009 & 0.134 & 0.171 & 0.238
& 0.638 & \textbf{0.958} & 0.321 & 0.483 
 \\
Fusionmamba                     
& 1.635 & \underline{0.974} & \underline{0.652} & 0.511
& 1.322 & 0.635 & \textbf{0.543} & 0.519
& 1.414 & 0.747 & 0.580 & 0.480
\\
TC-MoA                     
& \underline{1.661} & 0.811 & 0.565 & \underline{0.515}
& 1.562 & 0.577 & 0.477 & 0.522
& 1.556 & 0.579 & 0.508 & 0.466
\\   
MLFuse                
& 1.520 & 0.753 & 0.519 & 0.498
& 1.595 & 0.629 & 0.527 & 0.545
& 1.600 & 0.592 & 0.460 & \underline{0.501}
\\ 
Up-Fusion             
& 1.584 & 0.813 & 0.602 & 0.502
& \textbf{1.718} & \underline{0.644} & 0.536 & 0.533
& 1.616 & 0.816 & \textbf{0.621} & 0.496
\\ 
\hline
Ours                     
& \textbf{1.663} & \textbf{0.983} & \textbf{0.655} & \textbf{0.526}
& \underline{1.708} & \textbf{0.651} & \underline{0.540} & \textbf{0.548}
& \textbf{1.766} & \underline{0.952} & \underline{0.619} & \textbf{0.515}
\\ \hline
\end{tabular}
}
\end{table*}

In the MIF task, we employed medical images from the Harvard Medical website, comprising pairs such as MRI-CT, MRI-PET, and MRI-SPECT~\cite{mu2023mif1, wei2024mif2}. For this task, we compared our method with various deep learning-based multi-modal fusion techniques, including EMFusion~\cite{xu2021emfusion}, MSRPAN~\cite{fu2021MSRPAN}, SwinFusion~\cite{ma2022swinfusion}, Zero~\cite{lahoud2019Zero}, U2Fusion~\cite{xu2020u2fusion}, CDDFuse~\cite{zhao2023cddfuse}, PSLPT~\cite{wang2024PSLPT}, Fusionmamba~\cite{FusionMamba}, TC-MoA~\cite{zhu2024tcmoa}, PRRGAN~\cite{huang2024generative}, and TFS-Diff~\cite{xu2024simultaneous}.
\begin{figure*}[!htb]
    \centering
    \includegraphics[width=0.9\textwidth]{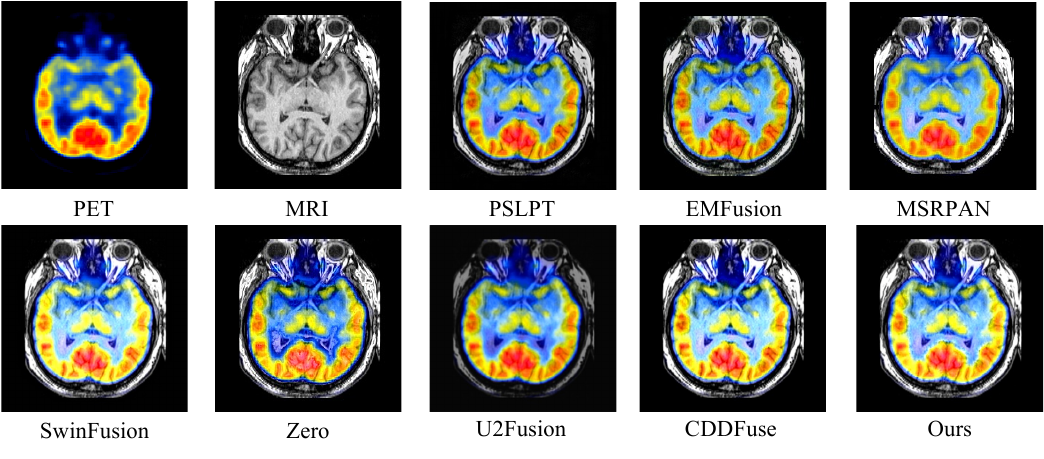}
    \caption{\label{result_PET}Comparative visual experiments of several methods on MRI-PET datasets}
    \label{result_PET}
\end{figure*}
\begin{figure*}[!htb]
    \centering
    \includegraphics[width=0.9\textwidth]{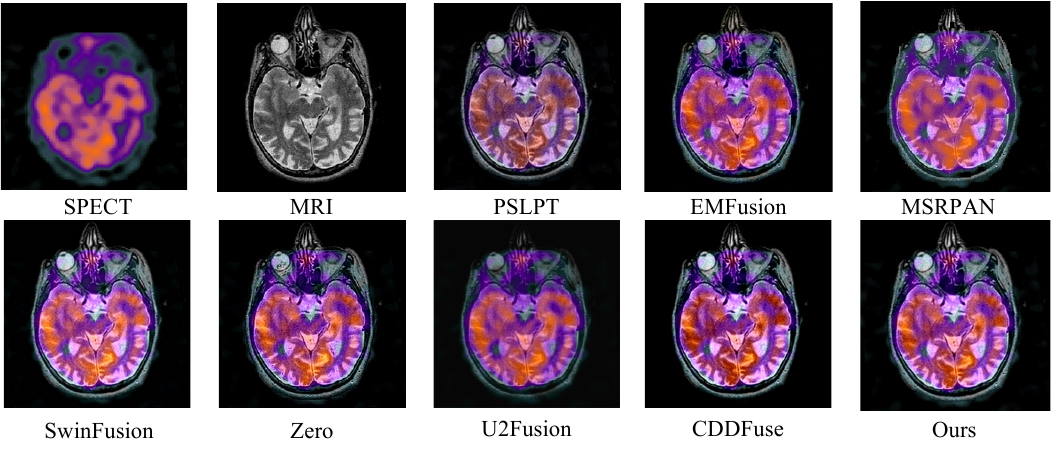}
    \caption{\label{result_SPECT}Comparative visual experiments of several methods on MRI-SPECT datasets}
    \label{result_SPECT}
\end{figure*}
\subsection{Implement Details}
All experiments were conducted using the PyTorch framework on two NVIDIA RTX 4090 GPUs. We trained for 500 epochs with a batch size of 4 for the pan-sharpening task and 200 epochs with a batch size of 1 for the medical image fusion (MIF) task. The network parameters were optimized using the Adam optimizer. The initial learning rate was set to $5e-4$, which was subsequently reduced to $5e-8$ using the CosineAnnealingLR scheduler over the specified epochs. For the pan-sharpening task, we randomly cropped training set images to obtain LRMS patches of size 32x32 and PAN images of size 128x128. For the MIF task, the training set images were cropped to 256x256.

\begin{figure*}[!htb]
    \centering
    \includegraphics[width=0.9\textwidth]{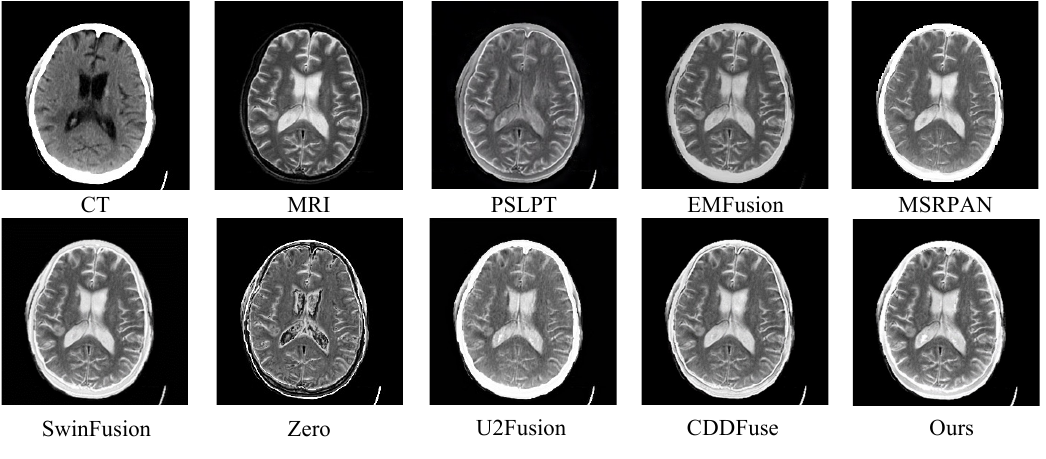}
    \caption{\label{result_CT}Comparative visual experiments of several methods on MRI-CT datasets}
    \label{result_CT}
\end{figure*}
\begin{table*}[!htb]
\caption{Ablation for our methods on three datasets. The PSNR/SSIM/SAM/ERGAS values on benchmarks are reported. The best results are shown in \textbf{bold}. RSS denotes Random-Shuffle Scanning.}
\label{table_abl}
\centering
\normalsize
 	\renewcommand{\tabcolsep}{3.0pt} % adjust horizontal space
\renewcommand{\arraystretch}{1.5}
\resizebox{\linewidth}{!}{
\begin{tabular}{c|c|c|c|c}
\hline
Ablation    & Variant     & WorldView-II    & Gaofen-2      & WorldView-III            \\ \hline
Baseline            & -      

& \textbf {42.3428}/\textbf {0.9734}/\textbf {0.0208}/\textbf {0.8840}
& \textbf {47.9180}/\textbf {0.9903}/\textbf {0.0097}/\textbf {0.5109} 
& \textbf {31.4005}/        {0.9327}/\textbf {0.0676}/\textbf {2.8098} 
\\ 
\hline
\multirow{3}{*}{Single Operation}
& (1) RM w/o RSS              
& 42.1460/0.9723/0.0211/0.9042 
& 47.4284/0.9890/0.0102/0.5355
& 31.3915/\textbf{0.9329}/0.0676/2.8113 
\\ 
& (2) RCIM w/o RSS       
& 42.2443/0.9729/0.0210/0.8919     
& 47.6847/0.9899/0.0099/0.5292
& 31.3476/0.9325/0.0680/2.8230
\\ 
& (3) RMIM w/o RSS
& 42.2136/0.9727/0.0211/0.8978 
& 47.5787/0.9895/0.0100/0.5324 
& 31.0971/0.9295/0.0704/2.9097 
\\ \hline
\multirow{4}{*}{Multi Operations}
& (1) + (2)
& 42.1205/0.9719/0.0214/0.9075
& 47.4000/0.9878/0.0104/0.5405
& 31.2150/0.9316/0.0692/2.8782
\\ 
& (1) + (3)
& 42.1128/0.9719/0.0215/0.9086
& 47.3851/0.9876/0.0105/0.5420
& 31.0547/0.9292/0.0706/2.9053
\\ 
& (2) + (3)
& 42.1684/0.9721/0.0213/0.9037 
& 47.4655/0.9889/0.0103/0.5381 
& 31.1200/0.9294/0.0702/2.9019
\\ 
& (1) + (2) + (3)
& 42.0765/0.9718/0.0218/0.9094 
& 47.3111/0.9857/0.0107/0.5498
& 31.0026/0.9274/0.0719/2.9142 
\\ \hline
\end{tabular}
}
\end{table*}

\begin{table}[!h]
\centering
\normalsize
\renewcommand{\arraystretch}{1.0}
\renewcommand{\tabcolsep}{3pt}
\caption{Ablation of different scanning strategies. The best results are shown in \textbf{bold}. RSS denotes Random-Shuffle Scanning, SS denotes Sequential Scanning, BD denotes Bidirectional Scanning, and DS denotes Diagonal Scanning \label{table_scanning}}
\resizebox{\linewidth}{!}
{
\begin{tabular}{c|cccc}
\hline
& \multicolumn{4}{c}{WorldView-II}                                          
\\ \cline{2-5} 
\multirow{-2}{*}{Scanning Strategy} 
& PSNR$\uparrow$      & SSIM$\uparrow$     & SAM$\downarrow$        & ERGAS$\downarrow$
\\ \hline

RSS -> SS
& 42.0967 &0.9718 &0.0218 &0.9094 
\\ 
RSS -> BS
& 42.2294 &0.9726 &0.0211 &0.8933 
\\ 
RSS -> DS
& 42.1539 &0.9721 &0.0214 &0.8967 
\\
\hline
Ours (RSS)                  
& \textbf {42.3428} &\textbf {0.9734} &\textbf {0.0208} &\textbf {0.8840}
\\ \hline
\end{tabular}
}
\end{table}

\subsection{Comparison with SOTA Methods}
\subsubsection{Pan-sharpening}
The experimental results on three datasets are presented in Table~\ref{table_pansharepening}. Reference metrics, including PSNR, SSIM, SAM, and ERGAS, are used to evaluate the fusion effect. The results demonstrate that the proposed method outperforms the SOTA methods across all metrics. Notably, in the PSNR metric, our method achieves improvements of 0.1047, 0.2727, and 0.2301 dB over Pan-mamba, which has the second-best performance. Better PSNR and SSIM evidence that the fusion results transfer more information from the original image and experience less distortion. In the qualitative comparison, Figure~\ref{result_WV3}, Figure~\ref{result_WV2}, and Figure~\ref{result_GF2} present representative samples from the respective datasets. Our method demonstrates superior performance in the mean squared error (MSE) graph, indicating that the fusion results are closer to the ground truth. Compared to other approaches, our method achieves more accurate restoration of both spectral and spatial details, underscoring the effectiveness of our fusion technique.

To further verify the generalization ability of our method in full-resolution scenes, we use three non-reference metrics, including ${{D}_{s}}$, ${{D}_{\lambda }}$, and QNR, to evaluate our method on the full WorldView-II dataset. The metric results are summarized in Table~\ref{table_fwv2}, while Figure~\ref{result_full} provides a visual comparison. Our method consistently outperforms the competing approaches across all metrics, highlighting the robust adaptability of Shuffle Mamba in image fusion tasks.

To comprehensively evaluate the parameter efficiency and computational performance of the proposed method, we further report additional metrics, including the number of parameters, GFLOPs, and actual inference time. All methods were benchmarked on $128 \times 128$ resolution images from the WorldView-II dataset, and the results are summarized in Table~\ref{table_flops}. Our method achieves a comparable parameter count to recent state-of-the-art (SOTA) approaches, while being significantly more lightweight—requiring only one-third to one-half the parameters of FAME~\cite{he2024frequencymoe} and DISPNet~\cite{DISPNet2024}. Moreover, it consistently outperforms the latest ARConv~\cite{wang2025arconv} method in terms of parameter count, computational cost, and inference time. As for training efficiency, due to substantial differences in model structures, we provide a focused comparison between Pan-mamba~\cite{he2024panmamba} and our method under the same number of epochs and settings. Pan-mamba completes training in 4.6301 hours (100\%), while our method takes 5.2194 hours (113\%). This indicates a modest 13\% increase in training time. However, our method achieves consistently superior performance across both quantitative and qualitative evaluations, demonstrating its effectiveness in enhancing image fusion quality. Overall, these results highlight the efficiency and practicality of the proposed Shuffle Mamba framework, offering a compelling trade-off between performance and computational cost.

\subsubsection{Medical Image Fusion}
The quantitative comparison results for four metrics on the MIF dataset are presented in Table~\ref{table_MIF}. The proposed method demonstrates strong performance across nearly all metrics, highlighting its effectiveness in medical image fusion. In our experimental results, a higher VIF score indicates closer alignment with human visual perception. Enhanced SCD, Qabf, and SSIM scores suggest that the fused image preserves greater similarity and undergoes less distortion compared to the original images. Qualitative comparisons of several methods on the MRI datasets are shown in Figure~\ref{result_PET}, Figure~\ref{result_SPECT}, and Figure~\ref{result_CT}. Our method consistently delivers superior visual quality, a finding further supported by the experimental metrics.
Furthermore, to evaluate perceptual quality, we performed a user study involving ten participants with medical backgrounds, including graduate students and clinical researchers. Each of the 72 MIF test pairs (ours vs. CDDFuse) was randomly displayed for pairwise comparison. The proposed method was preferred in 60 out of 72 cases (83.3\%), primarily due to clearer anatomical boundaries and improved soft-tissue visibility, supporting the quantitative superiority with perceptual evidence.

\subsubsection{Infrared and Visible Image Fusion}
To further evaluate the generalization capability of the proposed method across different multi-modal image fusion tasks, we additionally conducted experiments on Infrared and Visible Image Fusion (IVIF). In the IVIF setting, all methods were trained on the MSRS dataset~\cite{tang2022msrs} and tested on three benchmark datasets: MSRS, RoadScene~\cite{xu2020roadsence}, and M3FD~\cite{liu2022m3fd}. Besides the comparison methods used in the MIF experiments, including U2Fusion~\cite{xu2020u2fusion}, SwinFusion~\cite{ma2022swinfusion}, CDDFuse~\cite{zhao2023cddfuse}, PSLPT~\cite{wang2024PSLPT}, FusionMamba~\cite{FusionMamba}, and TC-MoA~\cite{zhu2024tcmoa}, we further extended the comparison to include SemLA~\cite{xie2023SemLA}, MLFuse~\cite{lei2025mlfuse}, and Up-Fusion~\cite{li2025text}.

Table \ref{table_IVIF} presents the quantitative results on three datasets. It can be observed that the proposed method consistently achieves the best overall performance across all evaluation metrics and datasets. Specifically, on the MSRS dataset, our method attains the highest scores in SCD, VIF, Qabf, and SSIM, indicating superior capability in preserving complementary modal information and structural consistency. For the RoadScene dataset, our method ranks first on three metrics and achieves the second-best performance on Qabf, demonstrating robust generalization to complex driving scenarios with varying illumination and scene dynamics. These results validate that the proposed framework not only excels in the pan-sharpening and MIF but also exhibits strong cross-modal generalization and adaptability in the IVIF task.

\subsection{Ablation Experiments}
To evaluate the effectiveness of the Random Shuffle mechanism, we conducted ablation experiments by removing the shuffle operation from the RM, RCIM, and RMIM modules, individually and in combination. As shown in Table~\ref{table_abl}, removing the shuffle operation consistently leads to performance degradation across all three datasets, confirming that the proposed scanning strategy is a key contributor to fusion quality. Notably, the degree of performance drop varies across datasets and modules. On WorldView-II, removing the shuffle in the RCIM module results in the largest decline. When multiple shuffle operations are removed simultaneously, the performance loss becomes more pronounced, demonstrating that these modules provide complementary gains. These observations confirm that the random-shuffle scanning strategy enhances global feature diversity and effectively reduces modality bias in different fusion stages.

In addition, we further evaluated the influence of different scanning strategies through a comparative ablation study, as presented in Table~\ref{table_scanning}. In this experiment, the default Random Shuffle Scanning (RSS) was replaced with Sequential Scanning (SS), Bidirectional Scanning (BS), and Diagonal Scanning (DS), respectively. The results show that RSS consistently outperforms all deterministic scanning strategies in all metrics. Specifically, while deterministic strategies such as BS and DS yield marginal improvements over SS, none of them can match the performance achieved by RSS. This confirms that introducing stochasticity through the Random Shuffle mechanism enables more diverse and global spatial interactions, effectively improves global context aggregation, and ultimately leads to superior fusion performance.

\begin{figure}[h]
% \captionsetup{type=figure}
\centering
\includegraphics[width=\linewidth]{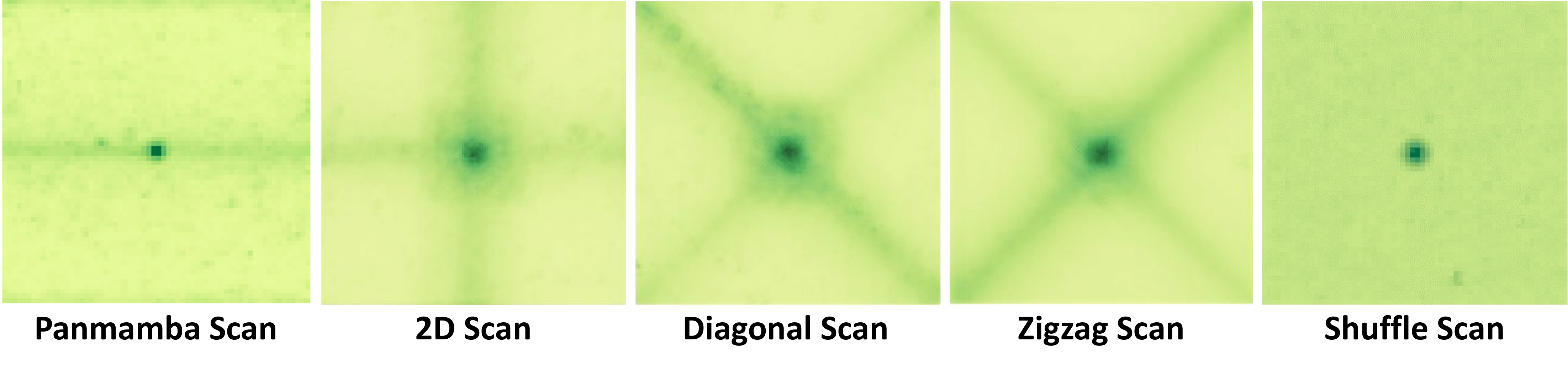}
\caption{\label{erf_scan}The effective receptive fields of different scanning schemes.
}
% \vspace{-0.5em}
\end{figure}

To validate the effectiveness of the proposed random shuffle scanning strategy, we provide a comparative visualization of the effective receptive fields (ERFs) under different scanning schemes. As illustrated in Figure~\ref{erf_scan}, the long-range dependencies captured by 2D scanning are primarily concentrated along horizontal and vertical directions. At the same time, diagonal and zigzag scans exhibit global receptive fields aligned with their respective diagonal paths. In contrast, our method produces a more uniformly distributed ERF, characterized by overall darker responses without prominent directional preferences. This demonstrates that the proposed random shuffle strategy effectively mitigates the local bias inherent in fixed scanning patterns, facilitating more balanced global modeling.
\begin{figure}[h]
% \captionsetup{type=figure}
\centering
\includegraphics[width=\linewidth]{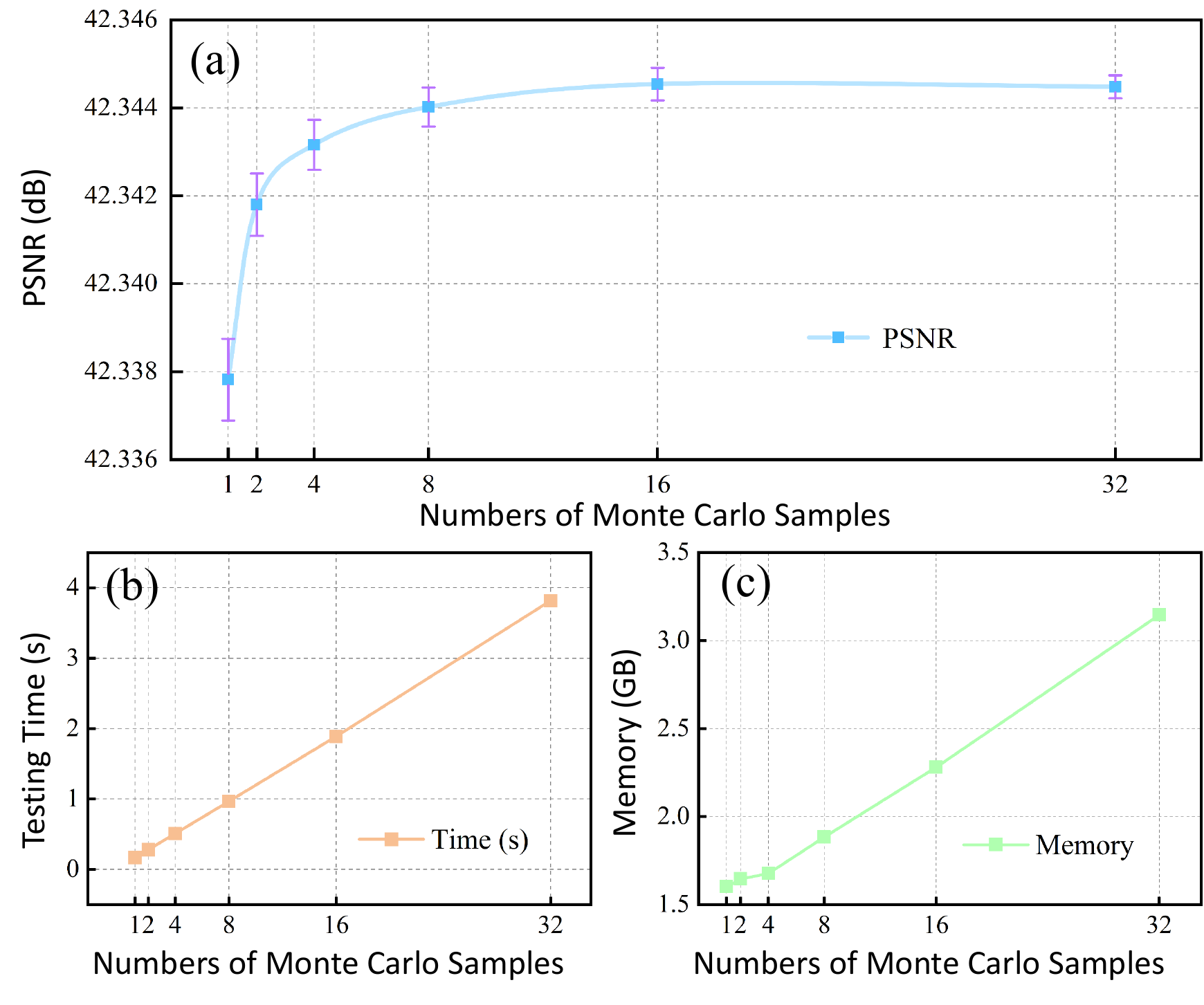}
\caption{\label{MC_image}The performance and costs in testing time and memory when choosing different numbers of samples.
}
% \vspace{-0.5em}
\end{figure}

Due to Monte Carlo averaging, Shuffle Mamba makes trading memory and computational resources possible for improved performance. We studied the relationship between the number of samples, the PSNR index, and resource consumption. We conducted five experiments for each sample size, calculating the average and standard deviation of various metrics. Figure~\ref{MC_image} (a) illustrates the PSNR trend, while Figure~\ref{MC_image} (b) and (c) detail the corresponding memory usage and the time required to process each image. As the number of samples increases, performance and resource consumption rise, enabling a trade-off between performance and efficiency. Additionally, Monte Carlo averaging enhances the theoretical robustness of the random shuffle operation, effectively improving the mean of PSNR while reducing the variance of the fusion result.
It is noteworthy that even when the number of Monte Carlo samples is set to 1, the network still performs well. This is because the overall architecture consists of multiple stacked modules, each with independent random shuffle scans. Therefore, multiple instances of random shuffling and recovery occur within the network, which further ensures stable performance and demonstrates the robustness of the proposed method.

\section{Conclusion} \label{s5}
In this paper, we replace the fixed scanning strategy used in current Mamba-based methods with a random shuffle-based scanning method and design a new Shuffle Mamba framework. This approach mitigates the bias introduced by fixed scanning methods and provides a global receptive field for multi-modal image fusion. During testing, we employ Monte Carlo averaging to account for the random factors introduced. Extensive experiments on two tasks demonstrate that our approach outperforms state-of-the-art methods and exhibits strong generalization capabilities.

Despite these advantages, our method has certain limitations. The primary drawback lies in the computational cost associated with the Monte-Carlo averaging during inference. As analyzed in Figure~\ref{MC_image}, while increasing the number of scan samples improves the results and reduces variance, it inevitably increases inference time and memory consumption linearly. Achieving the theoretical upper bound of performance requires multiple forward passes, which may limit real-time applications on resource-constrained edge devices. In future work, we aim to explore more efficient scanning strategies that maintain unbiased global perception without requiring repetitive sampling. Additionally, inspired by recent trends, we plan to extend the Shuffle Mamba framework to handle complex fusion scenarios, such as misaligned inputs and extreme weather conditions.

\bibliographystyle{IEEEtran}
\bibliography{cas-refs}

@article{jie2025fs,
  title={FS-Diff: Semantic guidance and clarity-aware simultaneous multimodal image fusion and super-resolution},
  author={Jie, Yuchan and Xu, Yushen and Li, Xiaosong and Zhou, Fuqiang and Lv, Jianming and Li, Huafeng},
  journal={Information Fusion},
  volume={121},
  pages={103146},
  year={2025},
  publisher={Elsevier}
}

@article{li2025umcfuse,
  title={UMCFuse: A Unified Multiple Complex Scenes Infrared and Visible Image Fusion Framework},
  author={Li, Xilai and Li, Xiaosong and Tan, Tianshu and Li, Huafeng and Ye, Tao},
  journal={IEEE Transactions on Image Processing},
  year={2025},
  publisher={IEEE}
}

@article{li2024allweather,
  title={All-weather multi-modality image fusion: Unified framework and 100k benchmark},
  author={Li, Xilai and Liu, Wuyang and Li, Xiaosong and Zhou, Fuqiang and Li, Huafeng and Nie, Feiping},
  journal={arXiv preprint arXiv:2402.02090},
  year={2024}
}

@article{li2025awmweather,
  title={AWM-Fuse: Multi-Modality Image Fusion for Adverse Weather via Global and Local Text Perception},
  author={Li, Xilai and Liu, Huichun and Li, Xiaosong and Ye, Tao and Kuang, Zhenyu and Li, Huafeng},
  journal={arXiv preprint arXiv:2508.16881},
  year={2025}
}

@article{li2025text,
  title={Text-Guided Channel Perturbation and Pretrained Knowledge Integration for Unified Multi-Modality Image Fusion},
  author={Li, Xilai and Li, Xiaosong and Jiang, Weijun},
  journal={arXiv preprint arXiv:2511.12432},
  year={2025}
}

@article{jie2024tsjnet,
  title={TSJNet: A multi-modality target and semantic awareness joint-driven image fusion network},
  author={Jie, Yuchan and Xu, Yushen and Li, Xiaosong and Tan, Haishu},
  journal={arXiv preprint arXiv:2402.01212},
  year={2024}
}

@article{huang2024generative,
  title={Generative adversarial network for trimodal medical image fusion using primitive relationship reasoning},
  author={Huang, Jingxue and Li, Xiaosong and Tan, Haishu and Cheng, Xiaoqi},
  journal={IEEE Journal of Biomedical and Health Informatics},
  year={2024},
  publisher={IEEE}
}

@inproceedings{xu2024simultaneous,
  title={Simultaneous tri-modal medical image fusion and super-resolution using conditional diffusion model},
  author={Xu, Yushen and Li, Xiaosong and Jie, Yuchan and Tan, Haishu},
  booktitle={International Conference on Medical Image Computing and Computer-Assisted Intervention},
  pages={635--645},
  year={2024},
  organization={Springer}
}

@article{tang2022msrs,
  title={Image fusion in the loop of high-level vision tasks: A semantic-aware real-time infrared and visible image fusion network},
  author={Tang, Linfeng and Yuan, Jiteng and Ma, Jiayi},
  journal={Information Fusion},
  volume={82},
  pages={28--42},
  year={2022},
  publisher={Elsevier}
}

@inproceedings{xu2020roadsence,
  title={Fusiondn: A unified densely connected network for image fusion},
  author={Xu, Han and Ma, Jiayi and Le, Zhuliang and Jiang, Junjun and Guo, Xiaojie},
  booktitle={Proceedings of the AAAI conference on artificial intelligence},
  volume={34},
  number={07},
  pages={12484--12491},
  year={2020}
}

@inproceedings{liu2022m3fd,
  title={Target-aware dual adversarial learning and a multi-scenario multi-modality benchmark to fuse infrared and visible for object detection},
  author={Liu, Jinyuan and Fan, Xin and Huang, Zhanbo and Wu, Guanyao and Liu, Risheng and Zhong, Wei and Luo, Zhongxuan},
  booktitle={Proceedings of the IEEE/CVF conference on computer vision and pattern recognition},
  pages={5802--5811},
  year={2022}
}

@article{xie2023SemLA,
  title={Semantics lead all: Towards unified image registration and fusion from a semantic perspective},
  author={Xie, Housheng and Zhang, Yukuan and Qiu, Junhui and Zhai, Xiangshuai and Liu, Xuedong and Yang, Yang and Zhao, Shan and Luo, Yongfang and Zhong, Jianbo},
  journal={Information Fusion},
  volume={98},
  pages={101835},
  year={2023},
  publisher={Elsevier}
}

@article{lei2025mlfuse,
  title={MLFuse: Multi-scenario feature joint learning for multi-modality image fusion},
  author={Lei, Jia and Li, Jiawei and Liu, Jinyuan and Wang, Bin and Zhou, Shihua and Zhang, Qiang and Wei, Xiaopeng and Kasabov, Nikola K},
  journal={IEEE Transactions on Multimedia},
  year={2025},
  publisher={IEEE}
}

@inproceedings{long2024dgmamba,
  title={Dgmamba: Domain generalization via generalized state space model},
  author={Long, Shaocong and Zhou, Qianyu and Li, Xiangtai and Lu, Xuequan and Ying, Chenhao and Luo, Yuan and Ma, Lizhuang and Yan, Shuicheng},
  booktitle={Proceedings of the 32nd ACM International Conference on Multimedia},
  pages={3607--3616},
  year={2024}
}

@article{lin2025eamamba,
  title={EAMamba: Efficient All-Around Vision State Space Model for Image Restoration},
  author={Lin, Yu-Cheng and Xu, Yu-Syuan and Chen, Hao-Wei and Kuo, Hsien-Kai and Lee, Chun-Yi},
  journal={arXiv preprint arXiv:2506.22246},
  year={2025}
}

@inproceedings{guo2024mambair,
  title={Mambair: A simple baseline for image restoration with state-space model},
  author={Guo, Hang and Li, Jinmin and Dai, Tao and Ouyang, Zhihao and Ren, Xudong and Xia, Shu-Tao},
  booktitle={European conference on computer vision},
  pages={222--241},
  year={2024},
  organization={Springer}
}

@article{shi2024msvmamba,
  title={Multi-scale vmamba: Hierarchy in hierarchy visual state space model},
  author={Shi, Yuheng and Dong, Minjing and Xu, Chang},
  journal={Advances in Neural Information Processing Systems},
  volume={37},
  pages={25687--25708},
  year={2024}
}

@inproceedings{wang2025arconv,
  title={Adaptive Rectangular Convolution for Remote Sensing Pansharpening},
  author={Wang, Xueyang and Zheng, Zhixin and Shao, Jiandong and Duan, Yule and Deng, Liang-Jian},
  booktitle={Proceedings of the Computer Vision and Pattern Recognition Conference},
  pages={17872--17881},
  year={2025}
}

@inproceedings{zhu2024tcmoa,
  title={Task-customized mixture of adapters for general image fusion},
  author={Zhu, Pengfei and Sun, Yang and Cao, Bing and Hu, Qinghua},
  booktitle={Proceedings of the IEEE/CVF conference on computer vision and pattern recognition},
  pages={7099--7108},
  year={2024}
}

@article{tcsvt_fusion1,
  title={Fusion2Void: Unsupervised Multi-focus Image Fusion Based on Image Inpainting},
  author={Lin, Huangxing and Lin, Yunlong and Xia, Jingyuan and Fan, Linyu and Li, Feifei and Wang, Yingying and Ding, Xinghao},
  journal={IEEE Transactions on Circuits and Systems for Video Technology},
  year={2024},
  publisher={IEEE}
}

@article{tcsvt_fusion2,
  title={An Infrared and Visible Image Fusion Method Based on Semantic-Sensitive Mask Selection and Bidirectional-Collaboration Region Fusion},
  author={Li, Xuan and Zhang, Guomin and Chen, Weiwei and Cheng, Li and Xie, Yining and Ma, Jiayi},
  journal={IEEE Transactions on Circuits and Systems for Video Technology},
  year={2024},
  publisher={IEEE}
}

@article{zhang2024frequency,
  title={Frequency decoupled domain-irrelevant feature learning for Pan-sharpening},
  author={Zhang, Jie and Cao, Ke and Yan, Keyu and Lin, Yunlong and He, Xuanhua and Wang, Yingying and Li, Rui and Xie, Chengjun and Zhang, Jun and Zhou, Man},
  journal={IEEE Transactions on Circuits and Systems for Video Technology},
  year={2024},
  publisher={IEEE}
}

@article{mu2023mif1,
  title={Learning to search a lightweight generalized network for medical image fusion},
  author={Mu, Pan and Wu, Guanyao and Liu, Jinyuan and Zhang, Yuduo and Fan, Xin and Liu, Risheng},
  journal={IEEE Transactions on Circuits and Systems for Video Technology},
  volume={34},
  number={7},
  pages={5921--5934},
  year={2023},
  publisher={IEEE}
}

@inproceedings{shahdoosti2012spatial,
  title={Spatial PCA as a new method for image fusion},
  author={Shahdoosti, Hamid Reza and Ghassemian, Hassan},
  booktitle={The 16th CSI International Symposium on Artificial Intelligence and Signal Processing (AISP 2012)},
  pages={090--094},
  year={2012},
  organization={IEEE}
}

@article{li2020laplacian,
  title={Laplacian redecomposition for multimodal medical image fusion},
  author={Li, Xiaoxiao and Guo, Xiaopeng and Han, Pengfei and Wang, Xiang and Li, Huaguang and Luo, Tao},
  journal={IEEE Transactions on Instrumentation and Measurement},
  volume={69},
  number={9},
  pages={6880--6890},
  year={2020},
  publisher={IEEE}
}

@inproceedings{veshki2020image,
  title={Image fusion using joint sparse representations and coupled dictionary learning},
  author={Veshki, Farshad G and Ouzir, Nora and Vorobyov, Sergiy A},
  booktitle={ICASSP 2020-2020 IEEE International Conference on Acoustics, Speech and Signal Processing (ICASSP)},
  pages={8344--8348},
  year={2020},
  organization={IEEE}
}

@inproceedings{liu2017medical,
  title={A medical image fusion method based on convolutional neural networks},
  author={Liu, Yu and Chen, Xun and Cheng, Juan and Peng, Hu},
  booktitle={2017 20th international conference on information fusion (Fusion)},
  pages={1--7},
  year={2017},
  organization={IEEE}
}

@article{xu2024pandenosing,
  title={Pan-denoising: Guided hyperspectral image denoising via weighted represent coefficient total variation},
  author={Xu, Shuang and Ke, Qiao and Peng, Jiangjun and Cao, Xiangyong and Zhao, Zixiang},
  journal={IEEE Transactions on Geoscience and Remote Sensing},
  year={2024},
  publisher={IEEE}
}

@article{bai2025refusion,
  title={Refusion: Learning image fusion from reconstruction with learnable loss via meta-learning},
  author={Bai, Haowen and Zhao, Zixiang and Zhang, Jiangshe and Wu, Yichen and Deng, Lilun and Cui, Yukun and Jiang, Baisong and Xu, Shuang},
  journal={International Journal of Computer Vision},
  volume={133},
  number={5},
  pages={2547--2567},
  year={2025},
  publisher={Springer}
}

@inproceedings{bai2025task,
  title={Task-driven Image Fusion with Learnable Fusion Loss},
  author={Bai, Haowen and Zhang, Jiangshe and Zhao, Zixiang and Wu, Yichen and Deng, Lilun and Cui, Yukun and Feng, Tao and Xu, Shuang},
  booktitle={Proceedings of the Computer Vision and Pattern Recognition Conference},
  pages={7457--7468},
  year={2025}
}

@article{xu2024hipandas,
  title={Hipandas: Hyperspectral Image Joint Denoising and Super-Resolution by Image Fusion with the Panchromatic Image},
  author={Xu, Shuang and Zhao, Zixiang and Bai, Haowen and Yu, Chang and Peng, Jiangjun and Cao, Xiangyong and Meng, Deyu},
  journal={arXiv preprint arXiv:2412.04201},
  year={2024}
}

@article{wei2024mif2,
  title={ECINFusion: A Novel Explicit Channel-wise Interaction Network for Unified Multi-modal Medical Image Fusion},
  author={Wei, Xinjian and Qiu, Yu and Xu, Xiaoxuan and Xu, Jing and Mei, Jie and Zhang, Jun},
  journal={IEEE Transactions on Circuits and Systems for Video Technology},
  year={2024},
  publisher={IEEE}
}

@article{xiao2023variational,
  title={Variational pansharpening based on coefficient estimation with nonlocal regression},
  author={Xiao, Jin-Liang and Huang, Ting-Zhu and Deng, Liang-Jian and Wu, Zhong-Cheng and Wu, Xiao and Vivone, Gemine},
  journal={IEEE Transactions on Geoscience and Remote Sensing},
  year={2023},
  publisher={IEEE}
}

@article{wu2023lrtcfpan,
  title={LRTCFPan: Low-rank tensor completion based framework for pansharpening},
  author={Wu, Zhong-Cheng and Huang, Ting-Zhu and Deng, Liang-Jian and Huang, Jie and Chanussot, Jocelyn and Vivone, Gemine},
  journal={IEEE Transactions on Image Processing},
  volume={32},
  pages={1640--1655},
  year={2023},
  publisher={IEEE}
}

@ARTICLE{FusionMamba,
  author={Peng, Siran and Zhu, Xiangyu and Deng, Haoyu and Deng, Liang-Jian and Lei, Zhen},
  journal={IEEE Transactions on Geoscience and Remote Sensing}, 
  title={FusionMamba: Efficient Remote Sensing Image Fusion With State Space Model}, 
  year={2024},
  volume={62},
  number={},
  pages={1-16},
  doi={10.1109/TGRS.2024.3496073}}

@article{wang2024PSLPT,
  title={A general image fusion framework using multi-task semi-supervised learning},
  author={Wang, Wu and Deng, Liang-Jian and Vivone, Gemine},
  journal={Information Fusion},
  volume={108},
  pages={102414},
  year={2024},
  publisher={Elsevier}
}

@article{xu2021emfusion,
  title={EMFusion: An unsupervised enhanced medical image fusion network},
  author={Xu, Han and Ma, Jiayi},
  journal={Information Fusion},
  volume={76},
  pages={177--186},
  year={2021},
  publisher={Elsevier}
}

@article{fu2021MSRPAN,
  title={A multiscale residual pyramid attention network for medical image fusion},
  author={Fu, Jun and Li, Weisheng and Du, Jiao and Huang, Yuping},
  journal={Biomedical Signal Processing and Control},
  volume={66},
  pages={102488},
  year={2021},
  publisher={Elsevier}
}

@inproceedings{lahoud2019Zero,
  title={Zero-learning fast medical image fusion},
  author={Lahoud, Fayez and S{\"u}sstrunk, Sabine},
  booktitle={2019 22th international conference on information fusion (FUSION)},
  pages={1--8},
  year={2019},
  organization={IEEE}
}

@article{ma2022swinfusion,
  title={SwinFusion: Cross-domain long-range learning for general image fusion via swin transformer},
  author={Ma, Jiayi and Tang, Linfeng and Fan, Fan and Huang, Jun and Mei, Xiaoguang and Ma, Yong},
  journal={IEEE/CAA Journal of Automatica Sinica},
  volume={9},
  number={7},
  pages={1200--1217},
  year={2022},
  publisher={IEEE}
}

@article{xu2020u2fusion,
  title={U2Fusion: A unified unsupervised image fusion network},
  author={Xu, Han and Ma, Jiayi and Jiang, Junjun and Guo, Xiaojie and Ling, Haibin},
  journal={IEEE Transactions on Pattern Analysis and Machine Intelligence},
  volume={44},
  number={1},
  pages={502--518},
  year={2020},
  publisher={IEEE}
}

@inproceedings{zhao2023cddfuse,
  title={Cddfuse: Correlation-driven dual-branch feature decomposition for multi-modality image fusion},
  author={Zhao, Zixiang and Bai, Haowen and Zhang, Jiangshe and Zhang, Yulun and Xu, Shuang and Lin, Zudi and Timofte, Radu and Van Gool, Luc},
  booktitle={Proceedings of the IEEE/CVF conference on computer vision and pattern recognition},
  pages={5906--5916},
  year={2023}
}

@inproceedings{he2024frequencymoe,
  title={Frequency-Adaptive Pan-Sharpening with Mixture of Experts},
  author={He, Xuanhua and Yan, Keyu and Li, Rui and Xie, Chengjun and Zhang, Jie and Zhou, Man},
  booktitle={Proceedings of the AAAI Conference on Artificial Intelligence},
  volume={38},
  number={3},
  pages={2121--2129},
  year={2024}
}

@inproceedings{DISPNet2024,
  title={Deep Unfolded Network with Intrinsic Supervision for Pan-Sharpening},
  author={Wang, Hebaixu and Gong, Meiqi and Mei, Xiaoguang and Zhang, Hao and Ma, Jiayi},
  booktitle={Proceedings of the AAAI Conference on Artificial Intelligence},
  volume={38},
  number={6},
  pages={5419--5426},
  year={2024}
}

@article{srivastava2014dropout,
  title={Dropout: a simple way to prevent neural networks from overfitting},
  author={Srivastava, Nitish and Hinton, Geoffrey and Krizhevsky, Alex and Sutskever, Ilya and Salakhutdinov, Ruslan},
  journal={The journal of machine learning research},
  volume={15},
  number={1},
  pages={1929--1958},
  year={2014},
  publisher={JMLR. org}
}

@article{he2024panmamba,
  title={Pan-Mamba: Effective pan-sharpening with State Space Model},
  author={He, Xuanhua and Cao, Ke and Yan, Keyu and Li, Rui and Xie, Chengjun and Zhang, Jie and Zhou, Man},
  journal={arXiv preprint arXiv:2402.12192},
  year={2024}
}

@inproceedings{xiao2023randomshuffle,
  title={Random shuffle transformer for image restoration},
  author={Xiao, Jie and Fu, Xueyang and Zhou, Man and Liu, Hongjian and Zha, Zheng-Jun},
  booktitle={International Conference on Machine Learning},
  pages={38039--38058},
  year={2023},
  organization={PMLR}
}

@article{ma2024umamba,
  title={U-mamba: Enhancing long-range dependency for biomedical image segmentation},
  author={Ma, Jun and Li, Feifei and Wang, Bo},
  journal={arXiv preprint arXiv:2401.04722},
  year={2024}
}

@article{zheng2024uvmamba,
  title={U-shaped Vision Mamba for Single Image Dehazing},
  author={Zheng, Zhuoran and Wu, Chen},
  journal={arXiv preprint arXiv:2402.04139},
  year={2024}
}

@article{huang2024localmamba,
  title={Localmamba: Visual state space model with windowed selective scan},
  author={Huang, Tao and Pei, Xiaohuan and You, Shan and Wang, Fei and Qian, Chen and Xu, Chang},
  journal={arXiv preprint arXiv:2403.09338},
  year={2024}
}

@misc{zhao2024rsmamba,
      title={RS-Mamba for Large Remote Sensing Image Dense Prediction}, 
      author={Sijie Zhao and Hao Chen and Xueliang Zhang and Pengfeng Xiao and Lei Bai and Wanli Ouyang},
      year={2024},
      eprint={2404.02668},
      archivePrefix={arXiv},
      primaryClass={cs.CV}
}

@article{chen2024rsmamba,
  title={RSMamba: Remote Sensing Image Classification with State Space Model},
  author={Chen, Keyan and Chen, Bowen and Liu, Chenyang and Li, Wenyuan and Zou, Zhengxia and Shi, Zhenwei},
  journal={arXiv preprint arXiv:2403.19654},
  year={2024}
}

@article{pnn,
	title={Pansharpening by convolutional neural networks},
	author={Masi, Giuseppe and Cozzolino, Davide and Verdoliva, Luisa and Scarpa, Giuseppe},
	journal={Remote Sensing},
	volume={8},
	number={7},
	pages={594},
	year={2016},
	publisher={Multidisciplinary Digital Publishing Institute}
}

@ARTICLE{srcnn,
  author={Dong, Chao and Loy, Chen Change and He, Kaiming and Tang, Xiaoou},
  journal={IEEE Transactions on Pattern Analysis and Machine Intelligence}, 
  title={Image Super-Resolution Using Deep Convolutional Networks}, 
  year={2016},
  volume={38},
  number={2},
  pages={295-307},
  doi={10.1109/TPAMI.2015.2439281}}

@article{IHS,
  title={Application of the IHS color transform to the processing of multisensor data and image enhancement},
  author={ Haydn, R.  and  Dalke, G. W.  and  Henkel, J.  and  Bare, J. E. },
  journal={National Academy of Sciences of the United States of America},
  volume={79},
  number={13},
  pages={571-577},
  year={1982},
}

@article{Brovey,
  title={Color enhancement of highly correlated images. II. Channel ratio and "chromaticity" transformation techniques - ScienceDirect},
  author={ Gillespie, A. R.  and  Kahle, A. B.  and  Walker, R. E. },
  journal={Remote Sensing of Environment},
  volume={22},
  number={ 3},
  pages={343-365},
  year={1987},
}

@inproceedings{GFPCA,
  title={Two-stage fusion of thermal hyperspectral and visible RGB image by PCA and guided filter},
  author={ Liao, W.  and  Xin, H.  and  Coillie, F. V.  and  Thoonen, G.  and  Philips, W. },
  booktitle={Workshop on Hyperspectral Image and Signal Processing: Evolution in Remote Sensing},
  year={2017},
}

@article{GS,
title={Process for Enhancing the Spatial Resolution of Multispectral Imagery Using Pan-Sharpening},
  author={ Laben, C.A. and Brower, B.V.},
  journal={US Patent 6011875A},
  year={2000},
}

@article{ATWT1999,
  title={Multiresolution-based image fusion with additive wavelet decomposition},
  author={Nunez, Jorge and Otazu, Xavier and Fors, Octavi and Prades, Albert and Pala, Vicenc and Arbiol, Roman},
  journal={IEEE Transactions on Geoscience and Remote sensing},
  volume={37},
  number={3},
  pages={1204--1211},
  year={1999},
  publisher={IEEE}
}

@article{HPF,
  title={Reconstruction of multispatial, multispectral image data using spatial frequency content},
  author={Schowengerdt, Robert A},
  journal={Photogrammetric Engineering and Remote Sensing},
  volume={46},
  number={10},
  pages={1325--1334},
  year={1980}
}

@article{fasbender2008bayesian,
	title={Bayesian data fusion for adaptable image pansharpening},
	author={Fasbender, Dominique and Radoux, Julien and Bogaert, Patrick},
	journal={IEEE Transactions on Geoscience and Remote Sensing},
	volume={46},
	number={6},
	pages={1847--1857},
	year={2008},
	publisher={IEEE}
}

@inproceedings{zhou2022innformer,
  title={Pan-sharpening with customized transformer and invertible neural network},
  author={Zhou, Man and Huang, Jie and Fang, Yanchi and Fu, Xueyang and Liu, Aiping},
  booktitle={Proceedings of the AAAI Conference on Artificial Intelligence},
  volume={36},
  number={3},
  pages={3553--3561},
  year={2022}
}

@inproceedings{pannet,
	title={PanNet: A deep network architecture for pan-sharpening},
	author={Yang, Junfeng and Fu, Xueyang and Hu, Yuwen and Huang, Yue and Ding, Xinghao and Paisley, John},
	booktitle={IEEE International Conference on Computer Vision},
	pages={5449--5457},
	year={2017}
}

@article{msdcnn,
  title={A Multiscale and Multidepth Convolutional Neural Network for Remote Sensing Imagery Pan-Sharpening},
  author={ Yuan, Q.  and  Wei, Y.  and  Meng, X.  and  Shen, H.  and  Zhang, L. },
  journal={IEEE Journal of Selected Topics in Applied Earth Observations and Remote Sensing},
  volume={11},
  number={3},
  pages={978-989},
  year={2018},
}

@article{tv,
  title={A new pansharpening algorithm based on total variation},
  author={Palsson, Frosti and Sveinsson, Johannes R and Ulfarsson, Magnus O},
  journal={IEEE Geoscience and Remote Sensing Letters},
  volume={11},
  number={1},
  pages={318--322},
  year={2013},
  publisher={IEEE}
}

@article{wald_gt,
author = {Wald, Lucien and Ranchin, Thierry and Mangolini, Marc},
year = {1997},
month = {11},
pages = {691-699},
title = {Fusion of satellite images of different spatial resolutions: Assessing the quality of resulting images},
volume = {63},
journal = {Photogrammetric Engineering and Remote Sensing}
}

@article{SFIM,
title={Smoothing filter-based intensity modulation:
A spectral preserve image fusion technique for improving
spatial details},
  author={ J. G. Liu.},
  journal={International Journal of Remote Sensing},
  volume={21},
  number={18},
  pages={3461-3472},
  year={2000},
}

@inproceedings{zhou2022sfiin,
  title={Spatial-frequency domain information integration for pan-sharpening},
  author={Zhou, Man and Huang, Jie and Yan, Keyu and Yu, Hu and Fu, Xueyang and Liu, Aiping and Wei, Xian and Zhao, Feng},
  booktitle={European Conference on Computer Vision},
  pages={274--291},
  year={2022},
  organization={Springer}
}

@article{he2023msddn,
  title={Multi-Scale Dual-Domain Guidance Network for Pan-sharpening},
  author={He, Xuanhua and Yan, Keyu and Zhang, Jie and Li, Rui and Xie, Chengjun and Zhou, Man and Hong, Danfeng},
  journal={IEEE Transactions on Geoscience and Remote Sensing},
  year={2023},
  publisher={IEEE}
}

@inproceedings{zhou2022panformer,
  title={Panformer: A transformer based model for pan-sharpening},
  author={Zhou, Huanyu and Liu, Qingjie and Wang, Yunhong},
  booktitle={2022 IEEE International Conference on Multimedia and Expo (ICME)},
  pages={1--6},
  year={2022},
  organization={IEEE}
}

@article{Mamba,
  title={Mamba: Linear-time sequence modeling with selective state spaces},
  author={Gu, Albert and Dao, Tri},
  journal={arXiv preprint arXiv:2312.00752},
  year={2023}
}

@article{s4,
  title={Efficiently modeling long sequences with structured state spaces},
  author={Gu, Albert and Goel, Karan and R{\'e}, Christopher},
  journal={arXiv preprint arXiv:2111.00396},
  year={2021}
}

@article{smith2022simplified,
  title={Simplified state space layers for sequence modeling},
  author={Smith, Jimmy TH and Warrington, Andrew and Linderman, Scott W},
  journal={arXiv preprint arXiv:2208.04933},
  year={2022}
}

@article{h3,
  title={Long range language modeling via gated state spaces},
  author={Mehta, Harsh and Gupta, Ankit and Cutkosky, Ashok and Neyshabur, Behnam},
  journal={arXiv preprint arXiv:2206.13947},
  year={2022}
}

@article{VisionMamba,
  title={Vision mamba: Efficient visual representation learning with bidirectional state space model},
  author={Zhu, Lianghui and Liao, Bencheng and Zhang, Qian and Wang, Xinlong and Liu, Wenyu and Wang, Xinggang},
  journal={arXiv preprint arXiv:2401.09417},
  year={2024}
}

@article{Vmamba,
  title={Vmamba: Visual state space model},
  author={Liu, Yue and Tian, Yunjie and Zhao, Yuzhong and Yu, Hongtian and Xie, Lingxi and Wang, Yaowei and Ye, Qixiang and Liu, Yunfan},
  journal={arXiv preprint arXiv:2401.10166},
  year={2024}
}

@inproceedings{yang2023panflownet,
  title={PanFlowNet: A Flow-Based Deep Network for Pan-sharpening},
  author={Yang, Gang and Cao, Xiangyong and Xiao, Wenzhe and Zhou, Man and Liu, Aiping and Chen, Xun and Meng, Deyu},
  booktitle={Proceedings of the IEEE/CVF International Conference on Computer Vision},
  pages={16857--16867},
  year={2023}
}

\end{document}